\documentclass{article}

\usepackage[main, final]{neurips_2026}

\usepackage[utf8]{inputenc}
\usepackage[T1]{fontenc}
\usepackage{hyperref}
\usepackage{url}
\usepackage{booktabs}
\usepackage{amsfonts}
\usepackage{amsmath,amssymb,amsthm,mathtools}
\usepackage{nicefrac}
\usepackage{microtype}
\usepackage{xcolor}
\usepackage{float}
\usepackage{enumitem}
\usepackage{tikz}
\usepackage{tcolorbox}
\usetikzlibrary{arrows.meta,calc,positioning,fit,backgrounds,shadows.blur}

\hypersetup{hidelinks,hypertexnames=false}
\usepackage{subcaption}
\usepackage{tabularx}
\usepackage{algorithm}
\usepackage{algpseudocode}
\usepackage{xcolor}

\usepackage{aliascnt}

\newtheorem{theorem}{Theorem}[section]

\newaliascnt{lemma}{theorem}
\newtheorem{lemma}[lemma]{Lemma}
\aliascntresetthe{lemma}

\newaliascnt{proposition}{theorem}
\newtheorem{proposition}[proposition]{Proposition}
\aliascntresetthe{proposition}

\newaliascnt{corollary}{theorem}
\newtheorem{corollary}[corollary]{Corollary}
\aliascntresetthe{corollary}

\newaliascnt{definition}{theorem}
\newtheorem{definition}[definition]{Definition}
\aliascntresetthe{definition}

\newaliascnt{assumption}{theorem}

\aliascntresetthe{assumption}

\newaliascnt{remark}{theorem}
\newtheorem{remark}[remark]{Remark}
\aliascntresetthe{remark}

\newaliascnt{example}{theorem}

\aliascntresetthe{example}

\usepackage[capitalize,noabbrev]{cleveref}
\makeatletter
\AddToHook{cmd/appendix/before}{\def\cref@section@alias{appendix}\def\cref@subsection@alias{appendix}}
\makeatother

\crefname{corollary}{Corollary}{corollaries}
\Crefname{corollary}{Corollary}{Corollaries}

\usepackage{tikz}
\usepackage{pgfplots}
\pgfplotsset{compat=1.18}

\definecolor{myOrange}{RGB}{235, 105, 92}
\definecolor{myMaroon}{RGB}{128, 0, 0}
\definecolor{maroon}{rgb}{0.5,0.0,0.0}
\definecolor{heavymaroon}{rgb}{0.3,0.0,0.0}

\definecolor{mygreen}{RGB}{160, 200, 140}

\hypersetup{ 
colorlinks=true, 
linkcolor=myOrange, 
citecolor=myMaroon, filecolor=cyan, 
urlcolor=mygreen
}

\usepackage{lipsum}                     %
\usepackage{xargs}
\usepackage{xfrac}
\usepackage{amssymb}
\usepackage{amsmath}
\usepackage{bm} 
\usepackage{mathtools} 
\usepackage{amsthm}
\usepackage{mleftright}
\usepackage{refcount}

\DeclareMathOperator{\poly}{poly}
\newcommand{\defeq}{\coloneqq}

\newcommand{\conv}{\mathrm{Conv}}

\def\[#1\]{\begin{align*}#1\end{align*}}

\NewDocumentCommand{\numberthis}{om}{%
  \IfNoValueTF{#1}{%
    \refstepcounter{equation}\tag{\theequation}%
  }{%
    \tag{#1}%
  }%
  \label{#2}%
}

\newcommand{\R}{\mathbb{R}}

\newcommand{\tr}{\operatorname{tr}}
\newcommand{\diag}{\operatorname{diag}}
\newcommand{\detm}{\operatorname{det}}
\newcommand{\Lip}{\operatorname{Lip}}
\newcommand{\Bn}{\operatorname{BN}}
\newcommand{\Id}{\mathbf{I}}
\newcommand{\Unif}{\operatorname{Unif}}
\newcommand{\dist}{\operatorname{dist}}
\newcommand{\rank}{\operatorname{rank}}

\newcommand{\cprod}{%
\mathop{\raisebox{-0.2ex}{\scalebox{1.6}{$\times$}}}\displaylimits
}

\newcommand{\one}{\mathbf{1}}

\newtheorem*{repeatthminner}{\repeatthmname}
\newcommand{\repeatthmname}{}

\newenvironment{repeatthm}[1]{%
  \renewcommand{\repeatthmname}{Theorem~\ref{#1}}%
  \begin{repeatthminner}%
}{%
  \end{repeatthminner}%
}

\newtheorem*{repeatlemmainner}{\repeatlemmaname}
\newcommand{\repeatlemmaname}{}

\newtheorem*{repeatpropinner}{\repeatpropname}
\newcommand{\repeatpropname}{}

\newenvironment{repeatprop}[1]{%
  \renewcommand{\repeatpropname}{Proposition~\ref{#1}}%
  \begin{repeatpropinner}%
}{%
  \end{repeatpropinner}%
}

\newtheorem*{repeatcorollaryinner}{\repeatcorollaryname}
\newcommand{\repeatcorollaryname}{}

\newenvironment{repeatcorollary}[1]{%
  \renewcommand{\repeatcorollaryname}{Corollary~\ref{#1}}%
  \begin{repeatcorollaryinner}%
}{%
  \end{repeatcorollaryinner}%
}

\title{On Lipschitz Explosion in Deep Neural Networks with Normalization: Consequences for Optimization
and Adversarial Robustness}

\newcommand{\NeurIPSAuthorEntry}[3]{%
  \begin{tabular}[c]{@{}c@{}}
    {\normalfont\bfseries #1}\\[0.12em]
    {\normalfont #2}\\[0.12em]
    {\normalfont\ttfamily #3}
  \end{tabular}%
}

\author{%
  \begin{tabular}{@{}c@{\hspace{1em}}c@{}}
    \NeurIPSAuthorEntry
      {Ashkan Soleymani}
      {Massachusetts Institute of Technology}
      {ashkanso@mit.edu}
    &
    \NeurIPSAuthorEntry
      {Reyhaneh Hosseinpourkhoshkbari}
      {The University of Texas at Dallas}
      {reyhaneh.hosseinpour@utdallas.edu}
    \\[2.5em]
    \NeurIPSAuthorEntry
      {Hadi Daneshmand}
      {University of Virginia}
      {dhadi@virginia.edu}
    &
    \NeurIPSAuthorEntry
      {Patrick Jaillet}
      {Massachusetts Institute of Technology}
      {jaillet@mit.edu}
  \end{tabular}%
}

\begin{document}

\maketitle

\begin{abstract}
    Can a network look stable at every layer and still be exponentially unstable as a whole?
    The Lipschitz constant of a neural network is a basic stability quantity appearing in optimization guarantees, generalization bounds, and robustness certificates. Deep networks naturally admit two such notions, \emph{input Lipschitzness}, measuring sensitivity to data perturbations, and \emph{parameter Lipschitzness}, measuring sensitivity to weight perturbations. We prove that, for deep networks with normalization layers, with batch normalization as the canonical example, \emph{both} Lipschitz constants can grow exponentially with depth and hence with parameter dimension, even under strong per-layer norm control such as $\|W_\ell\|_2\le 1$, and already for linear activations. Thus, a mechanism widely used to stabilize training can create severe worst-case instabilities that are invisible locally, i.e., from layerwise norm bounds alone. Parameter-Lipschitz explosion turns Lipschitz-dependent nonsmooth optimization guarantees into exponential-in-dimension bounds for deep normalized networks, while input-Lipschitz explosion makes worst-case Lipschitz-based generalization and robustness certificates vacuous. The latter also yields a concrete rank-separation mechanism for adversarial vulnerability. The theory predicts that perturbations introducing new singular directions should be amplified much more strongly than perturbations at the same noise level that remain within the input's original singular subspace. Experiments on MNIST, Fashion-MNIST, and CIFAR-10 support this prediction, showing that rank-creating perturbations cause substantially sharper drops in accuracy, confidence, and margins than same-subspace perturbations.
\end{abstract}

\section{Introduction} \label{sec:intro}

Increasing the depth of neural networks is a key driver of modern performance, boosting image classification~\citep{he2016deep} and text generation in large language models~\citep{radford2019language}. Yet increased depth also makes optimization more challenging, motivating architectural stabilizers such as residual connections~\citep{he2016deep} and normalization layers~\citep{ioffe2015batch}, which are now standard components of modern architectures. In this work, we take a closer look at  batch normalization in particular, a seemingly routine component of modern deep neural networks, and study its \emph{unexpected implications} for optimization, generalization, and adversarial robustness.

A recurring quantity linking depth to both optimization and stability is a model's (local) \emph{Lipschitz constant}. Crucially, deep networks come with two different worst-case Lipschitz notions, depending on what is perturbed. Writing $F_\theta$ for the network map with parameters $\theta$ (and using Frobenius / Euclidean norms for concreteness), we consider
\[
\Lip_{X}(F_\theta)
:= \sup_{X\neq Y}\frac{\|F_\theta(X)-F_\theta(Y)\|_F}{\|X-Y\|_F},
\qquad
\Lip_{\Theta}(F;X_{\rm in})
:= \sup_{\theta\neq \theta'}\frac{\|F_\theta(X_{\rm in})-F_{\theta'}(X_{\rm in})\|_F}{\|\theta-\theta'\|_2}.
\]
The \emph{parameter} Lipschitz constant $\Lip_\Theta$ governs how sharply the network output (and hence the induced training objective) can change as the weights move, and it therefore enters worst-case iteration-complexity guarantees in nonsmooth optimization. The \emph{input} Lipschitz constant $\Lip_X$ governs worst-case sensitivity to data perturbations, and it appears (often via upper bounds) in several generalization and robustness analyses based on margins and stability, including spectrally-normalized bounds and related Lipschitz controls~\citep{bartlett2017spectrally,bubeck2021universal}. We use these \emph{doubly Lipschitz} notions as a lens to study unexpected implications of normalization layers in optimization, generalization, and adversarial robustness.

\textbf{Main result: both Lipschitz constants can be exponentially large.}
We show that for deep networks with normalization layers (with batch normalization as the canonical example), \emph{both} $\Lip_X(F_\theta)$ and $\Lip_\Theta(F;X_{\rm in})$ can grow \emph{exponentially} with depth $m$, and consequently  exponentially with the ambient parameter dimension $d$. This occurs even under strong per-layer norm control (e.g., $\|W_\ell\|_2\le 1$, and in the input-Lipschitz construction even $W_\ell$ orthogonal), and already for linear activations. Put differently, we find that while normalization stabilizes many practical training pipelines~\citep{ioffe2015batch,salimans2016weight,ba2016layer,ioffe2017batch,daneshmand2021batch,lange2022batch}, \textit{it causes worst-case doubly Lipschitz explosion as depth increases}.

\vspace{-0.05cm}
\subsection{Implications of parameter-Lipschitz explosion for nonsmooth optimization}
\vspace{-0.05cm}

For smooth objectives, first-order methods admit dimension-independent rates for reaching approximate stationarity~\citep{nesterov2013introductory}. Neural-network training objectives, however, are typically \emph{nonsmooth} (e.g., due to ReLU activations, pooling layers or normalizations), so stationarity is captured by generalized-gradient notions. A classical tractable notion is \emph{Goldstein stationarity}~\citep{goldstein1977optimization}, and recent algorithms can find approximate Goldstein stationary points with oracle complexity polynomial in the Lipschitz constant $L$ of the objective~\citep{zhang2020complexity,davis2022gradient,jordan2023deterministic}. In our setting, $L$ is not an external constant; instead, it is induced by the network and our lower bound shows it can be $\exp(\Omega(d))$ for deep normalized architectures. Consequently, Lipschitz-based ``dimension-free'' guarantees of the form $\poly(L,1/\varepsilon,1/\delta)$ can unfortunately translate into \emph{exponential dependence on parameter $d$} for this model class. Moreover, lower bounds in nonsmooth optimization show that some polynomial dependence on $L$ is unavoidable in the worst case~\citep{kornowski2022complexity}, underscoring that one cannot expect a uniform theory for Goldstein stationarity that is both fully general and only polylogarithmic in $L$. Motivated by the optimization barrier created by exploding Lipschitz constants, we consider a relaxed directional notion (probabilistic Goldstein stationarity) and give a basic algorithm that finds such points with oracle complexity that does \emph{not} depend polynomially on $L$. This begets a new quest to find more refined solution concepts (with probabilistic Goldstein stationarity as a basic starting point) that are theoretically tractable in $\poly(d, 1/\varepsilon)$ time and hence in $\poly(d, 1/\varepsilon, \log L)$ time.

\vspace{-0.05cm}
\subsection{Implications of input-Lipschitz explosion for generalization and adversarial robustness}
\vspace{-0.05cm}

The same phenomenon impacts learning-theoretic uses of Lipschitz control. Since $\Lip_X(F_\theta)$ (or upper bounds on it) enters several margin-based generalization bounds and Lipschitz-based stability arguments~\citep{bartlett2017spectrally,bubeck2021universal,hou2023instance}, an exponential lower bound indicates that these worst-case guarantees can deteriorate rapidly with depth, even when each layer is individually norm-controlled. Therefore, we conclude that the Lipschitzness assumption may not be realistic for studying generalization in deep neural networks with normalization layers, since this assumption becomes vacuous exponentially fast with depth.

The input Lipschitzness $\Lip_X(F_\theta)$ lens on batch normalization also sheds new light on recent empirical evidence that 
batch normalization can have an outsized effect on adversarial robustness~\citep{galloway2019batch,wang2022removing,zeise2025batchnorm}. By showing theoretically that the Lipschitzness of deep neural networks grows exponentially fast with depth as a byproduct of batch normalization, we show that normalization layers significantly increase the sensitivity of neural-network outputs to noise in their inputs. Therefore, we can partially explain this empirical observation.

\begin{figure}[!t]
\centering
\resizebox{0.92\linewidth}{!}{%
\begin{tikzpicture}[
    image/.style={
        inner sep=0pt,
        outer sep=0pt
    },
    label/.style={
        font=\small,
        align=center
    },
    pred/.style={
        rectangle,
        rounded corners=8pt,
        align=left,
        font=\small\sffamily,
        inner sep=7pt,
        line width=1.1pt,
        blur shadow={shadow blur steps=5, shadow opacity=15}
    },
    fail/.style={
        pred,
        draw=red!40,
        fill=red!5
    },
    success/.style={
        pred,
        draw=green!40,
        fill=green!5
    },
    arrow/.style={
        ->,
        >=Stealth,
        draw=gray!60,
        line width=1.1pt,
        shorten >=3pt,
        shorten <=3pt
    }
]

\node[image] (orig) at (0,0) {
    \includegraphics[width=1.9cm,height=1.9cm,keepaspectratio]{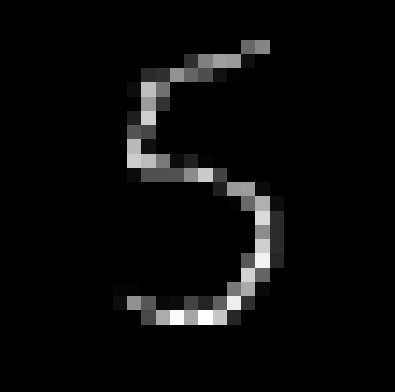}
};
\node[label, below=2pt of orig] {``Original Image''\\ True Label: 5};

\node[image] (iid-noise) at (3.5,1.0) {
    \includegraphics[width=1.55cm,height=1.55cm,keepaspectratio]{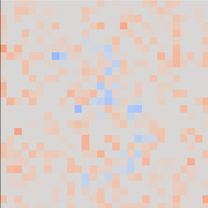}
};
\node[label, above=2pt of iid-noise] {Pixelwise Gaussian Noise};

\node[image] (iid-img) at (6.2,1.0) {
    \includegraphics[width=1.9cm,height=1.9cm,keepaspectratio]{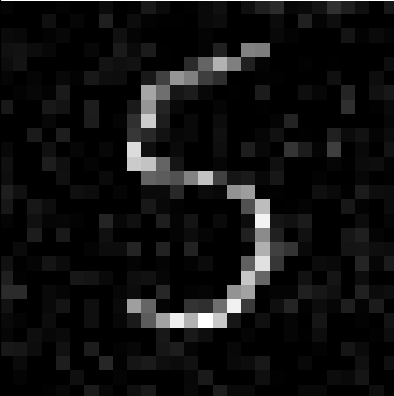}
};

\node[fail] (iid-pred) at (9.25,1.0) {
    Prediction: 0\\
    Confidence: 0.71
};

\node[image] (sub-noise) at (3.5,-1.0) {
    \includegraphics[width=1.55cm,height=1.55cm,keepaspectratio]{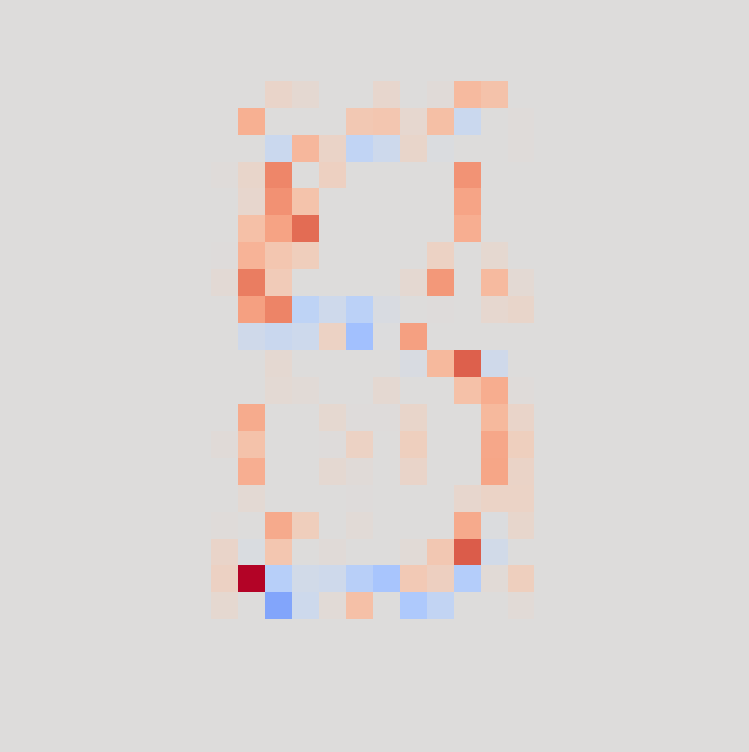}
};
\node[label, below=2pt of sub-noise] {Subspace Gaussian Noise};

\node[image] (sub-img) at (6.2,-1.0) {
    \includegraphics[width=1.9cm,height=1.9cm,keepaspectratio]{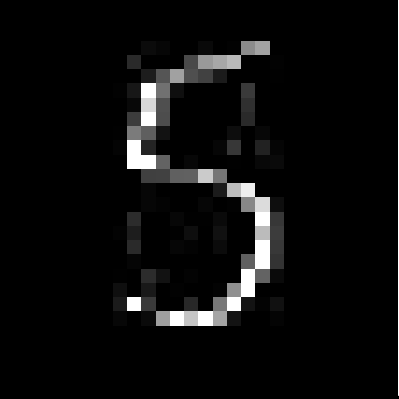}
};

\node[success] (sub-pred) at (9.25,-1.0) {
    Prediction: 5\\
    Confidence: 0.99
};

\node[image] (human) at (11.9,0) {
    \includegraphics[width=1.1cm]{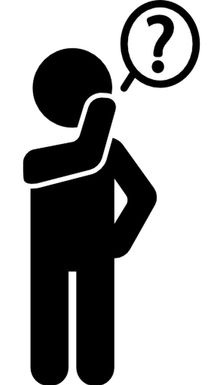}
};

\draw[arrow]
    (orig.east)
    to[out=28, in=180]
    node[pos=0.36, fill=white, inner sep=1pt, text=gray!70] {\Large $+$}
    (iid-noise.west);

\draw[arrow]
    (orig.east)
    to[out=-28, in=180]
    node[pos=0.36, fill=white, inner sep=1pt, text=gray!70] {\Large $+$}
    (sub-noise.west);

\draw[arrow] (iid-noise.east) -- (iid-img.west);
\draw[arrow] (iid-img.east) -- (iid-pred.west);

\draw[arrow] (sub-noise.east) -- (sub-img.west);
\draw[arrow] (sub-img.east) -- (sub-pred.west);

\draw[arrow] (iid-pred.east) to[out=0, in=150] ($(human.west)+(0,0.22cm)$);
\draw[arrow] (sub-pred.east) to[out=0, in=210] ($(human.west)+(0,-0.22cm)$);

\end{tikzpicture}%
}
\caption{We study the effect of different noise structures on the robustness of a multilayer neural network with normalization layers. Above, we add standard i.i.d.\ Gaussian noise; below, we instead apply noise restricted to the subspace of the original image. In the noisy images, red indicates positive values and blue indicates negative values. As depicted, even though the structured noise in the same subspace as the original image changes the image much more meaningfully (here, from a human perspective, making the 5 look more like an 8 as also observed in the noise pattern) the model still correctly classifies the image with high confidence. In contrast, under pixel-wise noise, despite the change being barely noticeable from a human perspective, the model is completely fooled and misclassifies the image with relatively high confidence. Our theoretical analysis predicts this phenomenon: the original image is low-rank, while i.i.d.\ Gaussian pixel noise makes the noisy image full-rank but nearly rank-deficient. Due to this relationship between the original image and the noisy one, normalization layers can amplify their separation across layers, causing neural networks' vulnerability to noise. This separation and vulnerability do not arise when Gaussian noise is added within the same subspace as the original image, even at a much higher magnitude.}
\label{fig:front}
\end{figure}

At a high level, our hardness construction for exponential growth of $\Lip_X(F_\theta)$ with depth exploits a rank-separation effect. The witness pair consists of a low-rank matrix and a nearby matrix that is full-rank but nearly rank-deficient. Although these two inputs are close in norm, normalization can amplify their separation across layers, leading to an exponential growth in the input Lipschitz constant. This mechanism suggests a corresponding empirical prediction for adversarial vulnerability under batch normalization: If a clean image, or a patch-based representation of it, has approximate low-rank structure (as suggested by classical low-rank patch models for natural images and by low-dimensional structure observed in datasets such as MNIST,  Fashion-MNIST, CIFAR-10, and ImageNet~\citep{savas2007handwritten,zhang2013learning,gu2014weighted,xu2015patch,liu2016content}) then adding small i.i.d. noise can generically make the representation full-rank while keeping it close to the low-rank set. Our theory predicts that batch-normalized networks amplify this small perturbation through depth, resulting in a large change in the output. We experimentally validate this prediction (see \Cref{fig:front}).

Our main contributions are as follows:
\begin{itemize}[leftmargin=1em, itemsep=1.5pt, topsep=0pt, parsep=3pt, partopsep=2pt]
    \item We distinguish two natural Lipschitz notions for deep networks: parameter Lipschitzness and input Lipschitzness. Through this lens, we reveal unexpected implications of normalization layers for optimization, generalization, and adversarial robustness.
    
    \item For deep networks with normalization layers, we show that both input and parameter Lipschitz constants can grow exponentially with depth (and hence with parameter dimension) even under strong per-layer norm constraints and even with linear activations.
    
    \item We clarify how parameter-Lipschitz explosion propagates to worst-case nonsmooth optimization. Existing guarantees for Goldstein stationarity that are polynomial in the objective Lipschitz constant $L$ become exponential in the parameter dimension for deep normalized networks. This motivates a new search for tractable stationarity notions with complexity polynomial in dimension and accuracy, but only polylogarithmic in $L$. We initiate this search by introducing probabilistic Goldstein stationarity, a basic directional relaxation of Goldstein stationarity derived from its dual formulation. We give a simple function-value algorithm that finds such points with oracle complexity that avoids polynomial dependence on $L$.

    \item We connect input-Lipschitz explosion to generalization and adversarial robustness. In particular, our results show that worst-case Lipschitz-based generalization, margin, and stability arguments can become vacuous exponentially fast with depth in normalized networks, even when each layer is norm-controlled. 

    \item More broadly, our construction identifies a simple mechanism for adversarial sensitivity in batch-normalized networks. Normalization can amplify small input differences when two nearby inputs have different rank structure. Consequently, when the data are approximately low rank, perturbations that introduce new singular directions can be magnified across layers, while perturbations that remain within the original singular subspace may be much less disruptive. Our experiments support this prediction across MNIST, Fashion-MNIST, and CIFAR-10. Perturbations that change the input's singular structure lead to substantially sharper drops in accuracy, confidence, and classification margins than perturbations at the same noise level that stay within the input's original singular subspace. This suggests a geometry-dependent failure mode of normalized networks and helps explain why batch normalization can have an outsized effect on adversarial robustness.
\end{itemize}

\section{Exponential Explosion of Two Lipschitz Constants with Dimension} \label{sec:exp_lower}

In this section, we study the Lipschitz constant of deep neural networks. In particular, we show that the presence of normalization layers, with a focus on batch normalization, causes the Lipschitz constant of the network to grow exponentially with depth, and consequently with the dimension of the parameter space~$d$. We observe this explosion in two notions of Lipschitz continuity: (i) the Lipschitz constant of the deep neural network with fixed weights, viewed as an input--output map with respect to the inputs, and (ii) the Lipschitz constant of the deep neural network with fixed input data, viewed as a function on the parameter space with respect to the weights.

\vspace{-0.1cm}
\subsection{Preliminaries and Notation}
\vspace{-0.1cm}

\paragraph{Notation.}
For a matrix $X$, $\|X\|_F$ is the Frobenius norm and $\|X\|_2$ the operator norm.
For a square matrix $A$, $\lambda_{\min}(A)$ denotes its smallest eigenvalue.
We write $X_{i:}$ for the $i$-th row of $X$ and $X_{:j}$ for the $j$-th column of $X$. We denote the set $\{1,2,\ldots,m\}$ by $[m]$.
\paragraph{Batch normalization (Column-wise normalization on $\R^{n\times k}$). }
Let $X\in\R^{n\times k}$. Define the diagonal matrix of column-squared norms
\[
D(X)\coloneqq \diag(X^\top X)\in\R^{k\times k},
\qquad
D(X)_{jj}=\|X_{:j}\|_2^2.
\]
When every column of $X$ is nonzero (equivalently $D(X)\succ 0$), the \emph{column-wise batch normalization}
map is $
\Bn(X)\coloneqq X\,D(X)^{-1/2}.$ Equivalently, $(\Bn(X))_{ij}=X_{ij}/\|X_{:j}\|_2$. We emphasize that the batch normalization operator defined above omits the mean-subtraction step for convenience, in line with prior formulations and theoretical studies~\citep{daneshmand2021batch,joudaki2023impact,meterez2024towards}. Our results extend to mean-subtracted batch normalization by an isometric embedding into the mean-zero subspace; \Cref{app:centering} gives the construction. If $D(X)\succ 0$, then every column of $\Bn(X)$ has unit norm, hence
\[
\numberthis{eq:fro_norm_fixed}
\|\Bn(X)\|_F^2 = \tr(\Bn(X)^\top\Bn(X))=k.
\]

\begin{remark}[Regularization at zero columns] \label{remark_esp_zero}
If one wants $\Bn$ to be defined on \emph{all} of $\R^{n\times k}$, one may work with the standard
$\varepsilon$-regularized variant
\[
\Bn_{\varepsilon}(X)\coloneqq X\,(D(X)+\varepsilon \Id_k)^{-1/2},
\qquad \varepsilon>0,
\]
and (when desired) take $\varepsilon\downarrow 0$ at the end.
Our lower bound proof below is written for $\Bn(X)=X\,D(X)^{-1/2}$ on the open set $\{D(X)\succ 0\}$,
and remains valid for $\Bn_\varepsilon$ with the same qualitative conclusion.
\end{remark}

\paragraph{A deep network with normalization layers and linear activation. }
Fix an integer $n\ge 1$, i.e., the batch size.
Let $k_0,k_1,\dots,k_m$ be positive integers (layer widths / feature dimensions).
For each layer $\ell\in\{1,\dots,m\}$ let $W_\ell\in\R^{k_{\ell-1}\times k_\ell}$ be a weight matrix. Given an input matrix $X^{(0)}\in\R^{n\times k_0}$, define the forward iterates
\begin{equation}\label{eq:network_recursion}
X^{(\ell)} \coloneqq f_{W_\ell}\!\bigl(X^{(\ell-1)}\bigr)
\;\;\stackrel{\mathrm{def}}{=}\;\;
\Bn\!\bigl(X^{(\ell-1)} W_\ell\bigr)
\in \R^{n\times k_\ell},
\qquad \ell=1,\dots,m,
\end{equation}
whenever the normalization is well-defined.
The depth-$m$ network map is
\[
F_m \coloneqq f_{W_m}\circ \cdots \circ f_{W_1}:\R^{n\times k_0}\to\R^{n\times k_m}.
\]
\paragraph{Why linear activations?}
We focus on linear activations because they provide a conservative setting for a lower bound. By removing nonlinearities, we rule out amplification effects that could come from the activation function itself, so any growth we prove must come from the interaction between the weights and normalization. This choice is also consistent with the mean-field analysis of batch-normalized networks in~\citet{yang2019mean}, where linear activations exhibit the slowest gradient-norm growth among the activation functions considered, even though the growth can still be exponential in depth. Thus, showing exponential Lipschitz growth already in the linear case demonstrates that the phenomenon is not caused by nonlinear activations. We expect the same mechanism to persist for a broader class of standard nonlinearities, we show this result for networks with ReLU activations in \Cref{app:relu} and leave the rest to future work.

\paragraph{Input and parameter Lipschitzness.}
We use two Lipschitz notions, depending on whether the perturbation is applied to the input or to the weights. First, fix a sequence of weights $(W_1,\dots,W_m)$ and and an input domain $\mathcal X$ on which every normalization step is defined. View the network as a map from inputs to outputs. Its input Lipschitz constant is
\[
\Lip_{\mathcal{X}}(F_m)\coloneqq 
\sup_{\substack{X,Y\in\mathcal X\\X\neq Y}}
\frac{\|F_m(X)-F_m(Y)\|_F}{\|X-Y\|_F}.
\]
Second, fix an input batch $X_{\mathrm{in}}\in\mathbb{R}^{n \times k_0}$ and view the network output as a function of the parameters. For a weight tuple $
\theta=(W_1,\dots,W_m) \in \Theta \subset \cprod_{\ell \in [m]} \mathbb{R}^{k_{\ell - 1} \times k_\ell}$,
we use the Euclidean norm obtained by vectorizing all weights, $
\|\theta\|_{2}\coloneqq 
\Big(\sum_{\ell=1}^m \|W_\ell\|_F^2\Big)^{1/2}$. The corresponding parameter-to-output map and its Lipschitz constant are
\[
\mathcal{F}_m(\theta)\coloneqq F_m^\theta(X_{\mathrm{in}}),
\qquad
\Lip_{\Theta}(\mathcal{F}_m)\coloneqq
\sup_{\substack{\theta,\theta'\in\Theta\\\theta\neq\theta'}}
\frac{\|\mathcal{F}_m(\theta)-\mathcal{F}_m(\theta')\|_F}
{\|\theta-\theta'\|_{2}}.
\]
The first quantity measures sensitivity to input perturbations, while the second measures sensitivity to changes in the weights.

\begin{remark}
    Normalization layer is invariant under positive rescaling since $\Bn(\alpha X)=\Bn(X)$ for every $\alpha>0$. Our exponential lower bounds are established on compact admissible
domains with scales bounded away from zero uniformly in depth.
Specifically, the input domains satisfy
$1\le \|X\|_F\le\sqrt{k}$, while every weight matrix in the
parameter witnesses has operator norm one.
\end{remark}

\vspace{-0.1cm}
\subsection{Explosion of the Input Lipschitz Constant $\Lip_{\mathcal{X}}(F_m)$}
\label{sec:main_lip_input}
\vspace{-0.1cm}

We first show that the input Lipschitz constant $\Lip_{\mathcal{X}}(F_m)$ can grow exponentially with depth. This growth is not an artifact of a large or expansive linear map. In fact, it already occurs in the following highly controlled setting with orthogonal weights:
\[
\numberthis{eq:square_setting}
n=k_0=k_1=\cdots=k_m=k\ge 2,\qquad X^{(0)}\in\R^{k\times k},\qquad W_\ell\in \mathbb{O}(k).
\]
Here each weight matrix is orthogonal $W_\ell \in \mathbb{O}(k)$, so $\|W_\ell\|_2=1$ and therefore $
\|XW_\ell\|_F\le \|X\|_F$. Thus, the linear part of every layer is non-expansive.

The setting is deliberately restrictive. The linear maps do not increase Frobenius norm, and by~\eqref{eq:fro_norm_fixed}, every post-normalization activation has fixed norm, $
\|X^{(\ell)}\|_F=\sqrt{k}$.
In particular, the activations remain in a bounded region throughout the network. Nevertheless, we show below that the input--output map can still have an input Lipschitz constant that grows exponentially with the depth $m$. Moreover, this growth is witnessed by uncountably many input pairs.

\begin{theorem}[Exponential lower bound on $\Lip_{\mathcal{X}}(F_m)$]
\label{thm:main}
Fix $k\ge 2$ and constant $\eta\in(0,1)$. There exist constants
$C_{k,\eta}>0$ and $\gamma_{k,\eta}>0$ such that, for every depth
$m\in\mathbb{N}$, one can choose orthogonal weights
$W_1,\dots,W_m\in\mathbb{O}(k)$ with $\|W_\ell\|_2=1$ for all
$\ell\in[m]$, for which the depth-$m$ map $F_m$ defined
in~\eqref{eq:network_recursion}--\eqref{eq:square_setting} satisfies $
\Lip_{\mathcal{X}}(F_m)\;\ge\; C_{k,\eta}\,e^{\gamma_{k,\eta} m}$,
where constants $C_{k,\eta}$ and $\gamma_{k,\eta}$ depend on $k,\eta$. Moreover, for this same fixed choice of weights, the lower bound is
witnessed by an uncountable family of input pairs.
\end{theorem}

Our proof construction uses \emph{orthogonal} weights, so the linear maps themselves do not create any expansion. The instability comes from the interaction between normalization and near rank-deficiency. We choose two inputs, $X_0$ and $Y_0$, that are exponentially close, with $Y_0$ rank-deficient and $X_0$ only slightly perturbed away from being rank-deficient. After $m$ carefully chosen normalization steps, the trajectory starting from $X_0$ becomes uniformly well-conditioned, while the trajectory starting from $Y_0$ remains rank-deficient. Thus the two outputs are separated by at least $\sqrt{\lambda_{\min}(X_m^\top X_m)}$ which is relatively high in a well-conditioned matrix, even though the original perturbation was exponentially small. This creates the exponential separation-to-perturbation ratio. We give the full proof of this result in \Cref{app:proof_main}, and provide a proof sketch in \Cref{sec:sketch}. 

The next corollary translates the depth-dependent lower bound into a lower bound in terms of the number of parameters. Thus, even when the parameter count is the measure of complexity, the input Lipschitz constant can still grow exponentially.

\begin{corollary}\label{cor:parambudget}
Fix $\eta\in(0,1)$ and set $k=2$.
There exist constants $C_\eta>0$ and $\Gamma_\eta>0$ such that, for every integer $d\ge 4$, there is a depth-$m$ network with at most $d$ scalar weight parameters whose input--output map satisfies $
\Lip_{\mathcal{X}}(F_m)\;\ge\; C_\eta\,e^{\Gamma_\eta d}$.
\end{corollary}

\begin{remark}
Although the constants in our general bounds in \Cref{thm:main} depend on $k$, the
exponential lower bounds hold uniformly over all widths $k\ge 2$.
Indeed, the width-two construction embeds into any larger width
via block-diagonal matrices, preserving the same exponential rate
and constants independently of $k$; see \Cref{app:width_embedding}.
\end{remark}

\vspace{-0.1cm}
\subsection{Explosion of the Parameter Lipschitz Constant $\Lip_{\Theta}(\mathcal{F}_m)$}
\label{sec:main_lip_param}
\vspace{-0.1cm}

In \Cref{sec:main_lip_input}, we showed that the input--output map can have an input Lipschitz constant $\Lip_{\mathcal{X}}(F_m)$ that grows exponentially with depth. We now show that the same phenomenon occurs when the input is fixed and the network is viewed as a function of its parameters. In other words, normalization can also lead to exponential growth of the parameter Lipschitz constant $\Lip_{\Theta}(\mathcal{F}_m)$.

\begin{theorem}[Exponential lower bound on $\Lip_{\Theta}(\mathcal{F}_m)$]
\label{thm:param_lip}
Fix $k\ge 2$ and constant $\eta\in(0,1)$. There exist constants
$\widetilde C_{k,\eta}>0$ and $\widetilde\gamma_{k,\eta}>0$ depending on $k$ and $\eta$ such that, for every depth $m\ge 2$, there is a depth-$m$ network of the form~\eqref{eq:network_recursion}, with
$\|W_\ell\|_2\le 1$ for all $\ell\in[m]$, for which the fixed-input parameter-to-output map $
\mathcal{F}_m(\theta)\coloneqq F_m^\theta(X_{\mathrm{in}})
$
with $X_{\mathrm{in}}=\Id_k$
satisfies $
\Lip_{\Theta}(\mathcal{F}_m)
\;\ge\;
\widetilde C_{k,\eta}\,e^{\widetilde\gamma_{k,\eta}m}.
$
\end{theorem}

The proof is a direct reduction from the input-Lipschitz construction in \Cref{thm:main}. We fix the input to be $X_{\mathrm{in}}=\Id_k$ and encode the two input witnesses from \Cref{thm:main} into the first-layer weights. The two parameter choices differ only in this first layer, while all subsequent weights are kept the same as in the input construction. By the scale-invariance of batch normalization, the resulting forward pass reproduces the same two trajectories after the first layer. Thus the parameter perturbation is as small as the original input perturbation, while the output separation remains bounded below, giving the same exponential separation ratio. Full details are provided in \Cref{app:exp_wrt_param}. As in \Cref{cor:parambudget}, this depth-dependent lower bound can also be translated into an exponential lower bound in the parameter dimension $d$.

\begin{corollary}\label{cor:parambudget_weights}
Fix any $\eta\in(0,1)$ and set $k=2$. There exist constants
$\widetilde C_\eta>0$ and $\widetilde \Gamma_\eta>0$ such that, for every integer $d\ge 8$, one can construct a depth-$m$ network with at most $d$ scalar weight parameters for which
$
\Lip_{\Theta}(\mathcal{F}_m)
\;\ge\;
\widetilde C_\eta\,e^{\widetilde \Gamma_\eta d}
$.
\end{corollary}

From the optimization viewpoint, the relevant quantity is the Lipschitz constant of the training objective that is fed by the network. To make this connection explicit, fix a realizable target parameter $\theta^\star\in\Theta$ and define
\[
f(\theta) \defeq \ \|\mathcal{F}_m(\theta)-\mathcal{F}_m(\theta^\star)\|_F, \qquad & \qquad \Lip(f) \defeq \sup_{\substack{\theta, \theta' \in \Theta \\
\theta \neq \theta'}} \frac{| f(\theta') - f(\theta) | }{ \| \theta' - \theta \|_2 },\numberthis{eq:def_objective}
\]
Here $\mathcal{F}_m(\theta^\star)$ plays the role of the observed label generated by the ground-truth network. The objective function $f(\theta)$ is equivalent to the mean-squared loss evaluated on the outputs of the neural network.

The next corollary shows that the exponential growth of $\Lip_{\Theta}(\mathcal{F}_m)$ transfers to the Lipschitz constant of this induced objective.

\begin{corollary}
\label{cor:objective_param_lip}
Let $\mathcal{F}_m$ be the parameter-to-output map constructed in \Cref{thm:param_lip}. There exists a realizable target parameter $\theta^\star\in\Theta$ such that the objective $f$ defined in~\eqref{eq:def_objective} satisfies
\[
\Lip(f)
\;\ge\;
\widetilde C_\eta e^{\widetilde \Gamma_\eta d},
\]
up to the constants in \Cref{cor:parambudget_weights}.
\end{corollary}

Thus, the same exponential sensitivity appears in the induced optimization objective and directly affects convergence guarantees that scale with its Lipschitz constant.

\vspace{-0.1cm}
\subsection{Experiments for Lipschitz Explosion}
\vspace{-0.1cm}

\begin{figure}[!t]
    \centering

    \begin{subfigure}{0.48\textwidth}
        \centering
        \includegraphics[width=\linewidth]{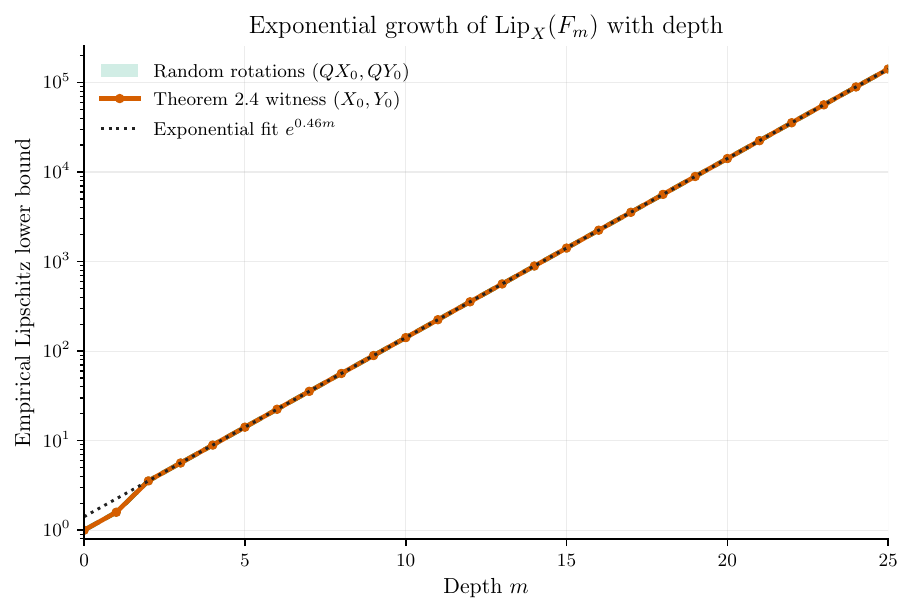}
        \caption{Constructed witnesses exhibit exponential growth.}
        \label{fig:main_image1}
    \end{subfigure}
    \hfill
    \begin{subfigure}{0.48\textwidth}
        \centering
        \includegraphics[width=\linewidth]{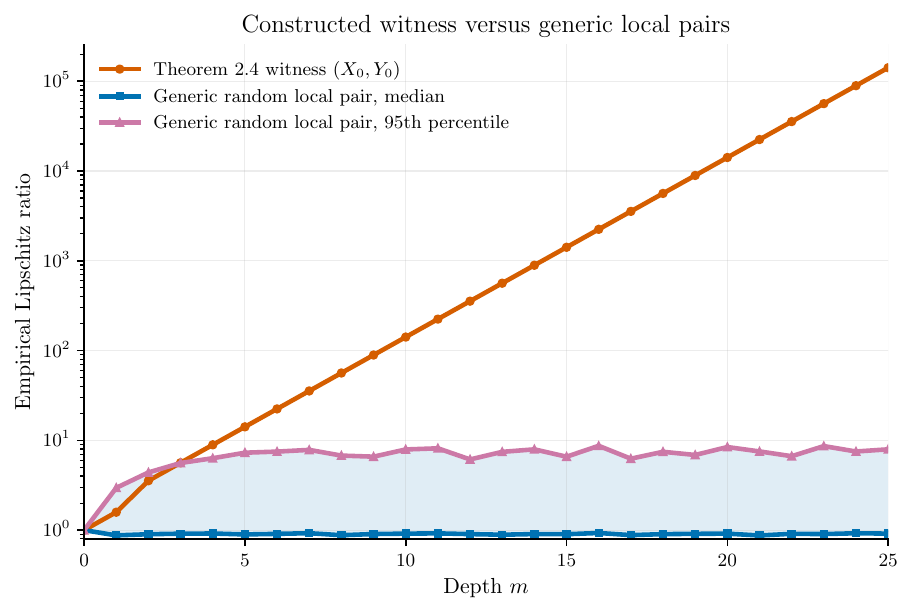}
        \caption{Generic local pairs do not exhibit the same growth.}
        \label{fig:main_image2}
    \end{subfigure}

    \caption{
    Empirical validation of Lipschitz explosion. 
    The constructed witness pair $(X_0,Y_0)$ and its random rotations $(QX_0,QY_0)$ grow exponentially with depth, while the median and $95$th percentile over generic random local pairs remain nearly flat.
    }
    \label{fig:lipschitz-experiments}
\end{figure}

We now empirically validate the exponential input-Lipschitz growth predicted by our theory. The results are shown in \Cref{fig:lipschitz-experiments} on a logarithmic scale. We focus on the setting of Corollary~\ref{cor:parambudget} with $k=2$. Since the proof of \Cref{thm:main} establishes existence through a nonconstructive probabilistic argument over the weights $\{W_\ell\}_{\ell=1}^m$, we use a constructive heuristic to select the weights in our experiments. Specifically, at each layer $\ell \in [m]$, we choose $W_\ell$ greedily so as to maximize the isotropy functional from Definition~\ref{def:isotropy}. In the two-dimensional setting, this amounts to a deterministic grid search over rotation angles. At each layer, we scan a grid of candidate rotations and select the one that maximizes the isotropy of the next activation. 

\Cref{fig:main_image1} compares the empirical Lipschitz ratio obtained from the constructed witness pair $(X_0,Y_0)$ with the predicted exponential growth. The same plot also shows witnesses obtained by applying random rotations $Q$ to the input witness pair, yielding pairs of the form $(QX_0,QY_0)$. As predicted by the orthogonal equivariance of the construction, these rotated witnesses exhibit the same exponential growth behavior. \Cref{fig:main_image2} compares the constructed witness with generic local input pairs. The random local pairs are generated from i.i.d. standard Gaussian perturbations, and we report both the median and the $95$th percentile of their empirical Lipschitz ratios. Unlike the constructed witness, these generic local pairs do not exhibit exponential growth with depth. These experiments support two complementary conclusions. First, the exponential growth predicted by our theory is clearly observable for the constructed witnesses. Second, this behavior is not typical for arbitrary local perturbations. Thus, the witness construction captures a genuine worst-case instability of normalized networks rather than a numerical artifact.

\vspace{-0.1cm}
\section{Implications for Optimization Theory in Deep Neural Networks}
\label{sec:imp_opt}
\vspace{-0.1cm}

We now turn to the optimization consequences of the lower bound
$L \defeq \Lip(f) \ge \exp \mleft( \Omega(d) \mright)$ from \Cref{sec:main_lip_param}.
Dimension-free convergence rates are often viewed as especially desirable in deep learning, since modern networks can have a very large number of parameters. Our results show that, for normalized networks, this interpretation can be misleading. A guarantee may be formally independent of the parameter dimension $d$, but if it depends polynomially on the objective Lipschitz constant $L$, then \Cref{cor:objective_param_lip} implies that the same guarantee can still scale exponentially with $d$. In this sense, Lipschitz-dependent ``dimension-free'' rates need not be truly dimension-free for deep normalized architectures.

This issue is particularly relevant in nonsmooth optimization. Neural-network training objectives are nonconvex, and they are often nonsmooth as well. Even if the outer loss is smooth, standard architectural components such as ReLU~\citep{nair2010rectified} and leaky-ReLU activations~\citep{maas2013rectifier}, max-pooling~\citep{lee2009convolutional}, absolute-value regularizers~\citep{scardapane2017group}, and piecewise-linear combinatorial layers~\citep{martins2016softmax,blondel2020fast} can introduce nonsmoothness into the objective. At the same time, these objectives are typically locally Lipschitz and differentiable almost everywhere, so generalized-gradient notions of stationarity provide the natural language for worst-case analysis. The relevant background on nonsmooth stationarity is recalled in \Cref{app:goldstein}; here, we focus on how the Lipschitz explosion above changes the interpretation of existing convergence guarantees.

\vspace{-0.1cm}
\subsection{Complexity of Finding Goldstein Stationary Points }
\label{sec:lower_bound_goldstein}
\vspace{-0.1cm}

Existing algorithms for finding $(\delta,\epsilon)$-approximate Goldstein stationary points have oracle complexity that depends polynomially on the objective Lipschitz constant $L$~\citep{zhang2020complexity,davis2022gradient,lin2022gradient,jordan2023deterministic,kong2023cost}. Such a dependence is acceptable when $L$ is viewed as a fixed property of the optimization problem. In deep normalized networks, however, $L$ is not an external constant; it is determined by the architecture and can grow with depth. Our results in \Cref{sec:exp_lower} show that $L$ can be as large as $\exp(\Omega(d))$. Consequently, a guarantee of the form $\mathcal{O}(\poly(L,1/\epsilon,1/\delta))$ can translate into exponential dependence on the parameter dimension $d$, even when the stated rate has no explicit dependence on $d$.

This raises a natural question: \textit{can one design algorithms for Goldstein stationarity whose dependence on $L$ is only logarithmic, for example with oracle complexity polynomial in $d$, $1/\epsilon$, $1/\delta$, and $\log L$?}

We show that, in full generality, the answer is negative. One cannot hope for a randomized first-order method whose oracle complexity is polynomial in $d$, $1/\epsilon$, $1/\delta$, and only logarithmic in $L$, uniformly over all Lipschitz objectives. The obstruction is not specific to neural networks or to nonconvexity: it already appears for convex Lipschitz functions. We rely directly on \citet[Theorem~3]{kornowski2022complexity}, which gives a lower bound on the oracle complexity of finding $(\delta,\epsilon)$-Goldstein stationary points of $1$-Lipschitz functions.

\begin{corollary}[Informal]
\label{cor:polyL-lb}
In the worst case, polynomial dependence on the Lipschitz constant $L$ is unavoidable for finding $(\delta,\epsilon)$-Goldstein stationary points. In particular, no randomized first-order algorithm can have oracle complexity only $\poly(d,1/\epsilon,1/\delta,\log L)$ uniformly over all Lipschitz objectives.
\end{corollary}

\vspace{-0.1cm}
\subsection{Toward Tractable Solution Concepts}
\label{sec:quest}
\vspace{-0.1cm}

These observations highlight a mismatch between existing nonsmooth optimization guarantees and the regimes that arise in deep normalized networks. As we observed, current guarantees for Goldstein stationarity do not yield truly polynomial-time bounds once the architecture-induced Lipschitz constant $L$ is taken into account. This suggests a natural quest: to identify weaker, but still meaningful, stationarity notions whose complexity is polynomial in $d$, $1/\varepsilon$, and $1/\delta$, and polynomial only in $\log L$. As a first step in this direction, we introduce probabilistic Goldstein stationarity, a simple directional relaxation of Goldstein stationarity. We use it as \emph{a basic baseline rather than a final answer}, and give a simple function-value algorithm that finds such points with number of queries independent of $L$.

Goldstein stationarity is a worst-case directional condition. To see this, fix $\delta>0$ and let $\partial_\delta f(x)$ denote the Goldstein subdifferential. For a direction $v\in\mathbb{S}^{d-1}$, consider the support function of the set of Goldstein subdifferentials $
m_\delta(x;v)\;\coloneqq\;\min_{g\in\partial_\delta f(x)}\langle g,v\rangle.$ By Lemma~\ref{lem:minimax_phi}, Goldstein stationarity can be written in the equivalent form
$
\dist(0,\partial_\delta f(x))\le \varepsilon
\Longleftrightarrow
\max_{v\in\mathbb{S}^{d-1}} m_\delta(x;v)\le \varepsilon$. Thus, a $(\delta,\varepsilon)$-Goldstein stationary point is one for which no direction has a large Goldstein margin. This dual viewpoint goes back to the work of~\citet{goldstein1977optimization}. We now relax this worst-case requirement by asking only that large-margin directions occupy a small fraction of the sphere.

\begin{definition}[Probabilistic Goldstein stationarity (PGS)]
\label{def:pgs}
Fix $\delta>0$, a margin threshold $\tau>0$, and a probability level $\rho\in(0,1)$.
Let $V\sim \Unif(\mathbb{S}^{d-1})$. We say that $x$ is \emph{$(\delta,\tau,\rho)$-PGS} if $
\Pr\big[m_\delta(x;V)\ge \tau\big]\ \le\ \rho$. We write $x$ is \emph{$(\delta,\varepsilon)$-PGS} as shorthand for $(\delta,\tau,\rho)=(\delta,\varepsilon,\varepsilon)$.
\end{definition}

We view probabilistic Goldstein stationarity (PGS) as a first step toward understanding which stationarity notions are tractable for deep normalized networks, rather than as a definitive solution concept. Additionally, we give a simple algorithm for finding PGS points that uses only function valuations and prove its convergence guarantee.

\begin{theorem}
\label{thm:pgd}
There exists a randomized function-value algorithm that finds a $(\delta,\tau,\rho)$-PGS point with probability at least $1-\gamma$ using $
\mathcal{O}\!\left(
\frac{\Delta}{\tau\delta\rho}
\log\frac{\Delta}{\gamma\tau\delta}
\right) $
queries, with no polynomial dependence on $L$.
\end{theorem}

\vspace{-0.1cm}
\section{Implications for Generalization and Robustness in Deep Neural Networks}
\label{sec:imp_gene_robust}
\vspace{-0.1cm}

The implications of Lipschitz explosion results for generalization are relatively direct. As discussed in \Cref{sec:intro}, blindly assuming Lipschitzness $L$ for deep normalized neural networks can make the resulting generalization bounds vacuous with depth (and number of parameters). We now turn to the unexpected implications of our results for the robustness of neural networks.

We next connect Lipschitz explosion to decision flips, showing that it can directly undermine the robustness of deep normalized networks to input noise. The input-Lipschitz lower bound in \Cref{thm:main} shows that an exponentially small input perturbation can create a large separation in the network output. One might suspect that this only makes global Lipschitz certificates overly pessimistic, without leading to an actual change in the predicted label. The next two informal consequences show that this is not the case. The same construction can be converted into an explicit binary robustness failure. We defer the formal statements and proofs to \Cref{app:adv_robustness}. We first show that a constant-margin decision flip can occur within an exponentially small adversarial radius. This provides a theoretical mechanism that helps explain why batch normalization can have an outsized effect on adversarial robustness~\citep{galloway2019batch,wang2022removing,zeise2025batchnorm}.

\begin{theorem}[Informal] \label{thm:adv_radius}
For every depth $m$, consider the same batch-normalized network as in \Cref{thm:main} and append a linear readout of Frobenius norm one. The resulting binary classifier has a clean input with score margin $\Omega_{k,\eta}(1)$, but the clean input can be misclassified after a perturbation of Frobenius norm at most $e^{-\Omega_{k,\eta}(m)}$.
\end{theorem}

\begin{figure}[!t]
    \centering
    \includegraphics[width=\linewidth]{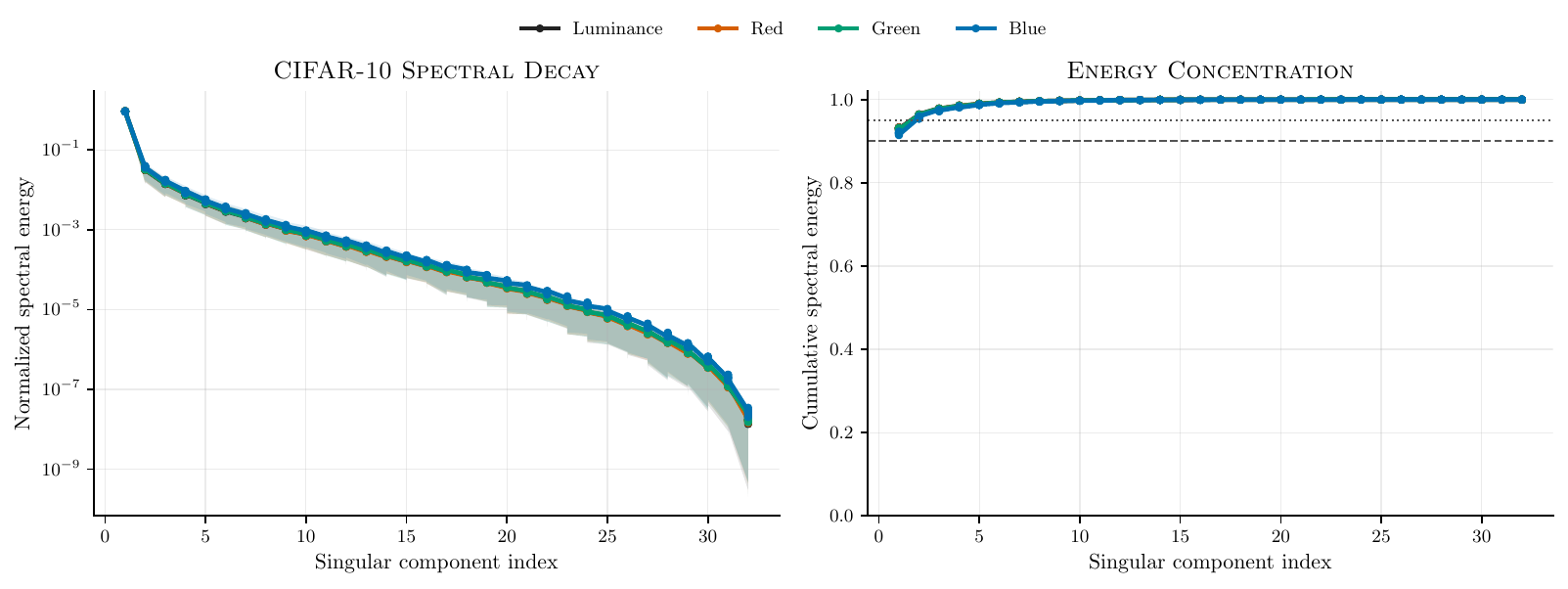}
    \caption{Spectral structure of CIFAR-10 luminance and RGB channel matrices. Left: normalized Gram eigenvalues $\lambda_j(X^\top X)/\tr(X^\top X)$. Right: cumulative spectral energy, with horizontal lines at $90\%$ and $95\%$. Most of the energy lies in the first few of the $32$ singular components.}
    \label{fig:cifar10_spectral_summary}
\end{figure}

The readout is the hyperplane placed halfway between the two output features generated by the witness pair in the proof of \Cref{thm:main}. Thus, the result does not rely on a separate instability mechanism. It shows that the Lipschitz explosion itself can produce a genuine small-radius adversarial example once the learned representation is passed through a linear classifier. We now explain the mechanism behind the separation observed in \Cref{fig:front}. The key point is that the adversarial perturbation is not only small in norm, but also rank-creating. This allows normalization to amplify the perturbation across layers, whereas perturbations that remain within the same low-rank branch do not experience the same amplification. This perspective is also consistent with the empirical geometry of standard image datasets, since MNIST, Fashion-MNIST, and CIFAR-10 images exhibit strong approximate low-rank structure in \Cref{fig:mnist_fashion_spectral_structure,fig:cifar10_spectral_summary}.

\begin{proposition}[Informal] \label{prop:rank_creating}
In the same construction, the clean input is rank-deficient, while the adversarially perturbed input is full-rank and only $e^{-\Omega_{k,\eta}(m)}$ away. Moreover, the full-rank adversarial output is a constant distance away from the entire rank-deficient output branch of the network. Thus perturbations that leave the low-rank branch and perturbations that stay within it have fundamentally different behavior under normalization.
\end{proposition}

\begin{figure}[!t]
    \centering
    \includegraphics[width=1.01\linewidth]{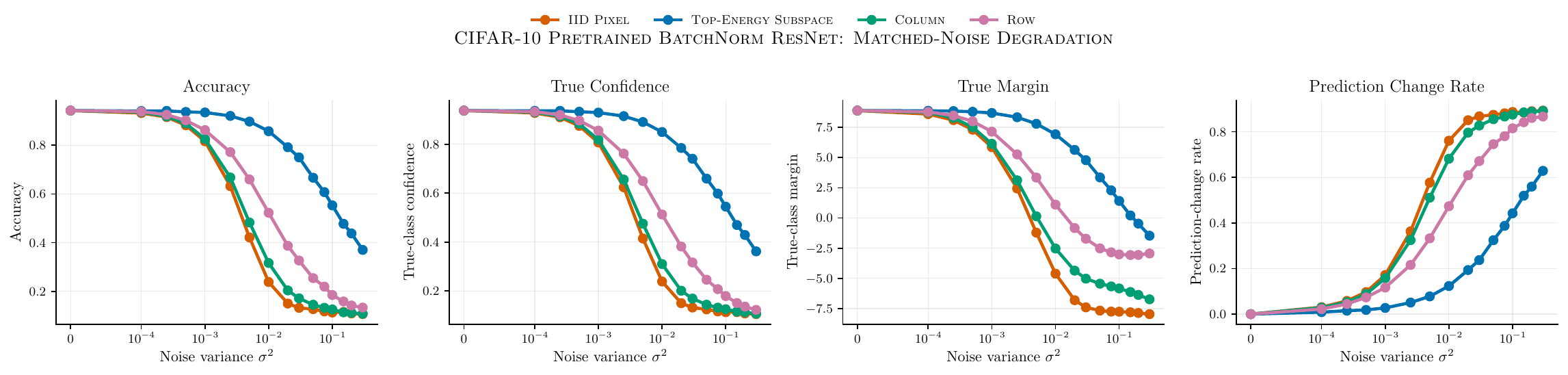}
    \caption{
    Effect of noise structure on a pretrained BatchNorm ResNet56 on CIFAR-10. Same-subspace perturbations are substantially less harmful than pixel-, row-, and column-wise noise.}
    \label{fig:cifar10_main}
\end{figure}

We empirically validate the prediction of \Cref{prop:rank_creating} in both realistic and controlled settings. In \Cref{fig:cifar10_main}, we use a ResNet56 model pretrained on CIFAR-10, while in \Cref{fig:matched_noise_degradation_all} we use toy CNNs with batch normalization trained on MNIST and Fashion-MNIST. In all experiments, we vary the noise variance and compare perturbations with matched pre-clipping energy but different geometric structure. We consider four perturbation families: i.i.d. pixel-wise noise, i.i.d. column-wise noise, i.i.d. row-wise noise, and singular-value noise, which perturbs only the singular values and therefore keeps the image in the same singular subspace.

The results are consistent with the mechanism suggested by \Cref{prop:rank_creating}. Perturbations that introduce new directions or alter the rank structure of the input lead to a much stronger degradation in accuracy, confidence, and margin. In contrast, same-subspace perturbations are substantially less effective, even at comparable noise energy. This supports the view that the vulnerability induced by normalization is not determined only by perturbation size, but also by whether the perturbation moves the input away from its underlying low-rank structure. We defer the detailed experimental setup and additional results to \Cref{app:exp}.

\section{Conclusion}

We showed that normalization layers can induce exponential growth in both input and parameter Lipschitz constants, even under strong per-layer norm control and already with linear activations. This Lipschitz explosion changes the interpretation of several standard guarantees for deep normalized networks, making Lipschitz-based optimization, generalization, and robustness analyses potentially vacuous in worst-case regimes. Our results also identify a rank-separation mechanism behind adversarial sensitivity and show that this mechanism is reflected empirically. We hope these findings motivate new stability notions and analysis tools that better capture the behavior of modern normalized architectures.

\section*{Acknowledgement}

The authors thank Suvrit Sra for introducing them to nonsmooth optimization and its literature.

A.S. and P.J. were partially supported by the ONR grant N00014-24-1-2470. Part of this research was performed while A.S. was visiting the Institute for Mathematical and Statistical Innovation (IMSI), which is supported by the National Science Foundation (Grant No. DMS-2425650).

\bibliographystyle{unsrtnat}
\bibliography{refs} 

\newpage
\appendix

\section{Related Works}
\label{app:additional_related_work}

\paragraph{Deep learning theory.}
Our study is motivated by the fact that depth is both powerful and difficult to manage. Deeper networks have driven many empirical gains, but they also tend to make optimization more fragile. Many standard tools in modern architectures were introduced to make very deep models easier to train, including normalization layers~\citep{ioffe2015batch,ba2016layer}, residual skip connections~\citep{he2016deep}, and modified activation functions~\citep{klambauer2017self}. In practice, these tools often stabilize training and make it possible to scale networks to much greater depths. We ask whether this practical stabilization can come with unanticipated costs for training, robustness, and statistical behavior. As a first step toward this broader question, in this work, we study batch normalization.

The Lipschitz constant of a neural network appears in several learning-theoretic analyses through Lipschitz- and margin-sensitive complexity measures. For example, \citet{bartlett2017spectrally} prove margin-based generalization bounds whose complexity term depends on a spectral complexity of the network, given by the product of the layer spectral norms together with an additional correction factor. Similarly, \citet{bubeck2021universal} study robustness through the Lipschitz constant, showing tradeoffs between interpolation, model size, and smoothness. These results make Lipschitz control a natural quantity to study when reasoning about generalization and robustness. Our contribution is complementary. We show that, in deep networks with normalization layers, the relevant input Lipschitz constant can itself grow exponentially with depth, even under strong per-layer norm control. Thus, Lipschitz-dependent generalization and robustness guarantees can become vacuous for deep normalized networks in the worst case.

Our analysis is closely related to mean-field studies of batch normalization. 
\citet{yang2019mean} show that, in the infinite-width limit, gradient norms in batch-normalized networks can grow exponentially with depth. 
More recently, \citet{meterez2024towards} showed that this explosion can be avoided for linear activations under suitable full-rank conditions on the input batch. 
Our results show that this conditioning requirement is essential. 
Even with linear activations and norm-controlled weights, inputs that are close to rank deficiency can lead to exponential growth of the worst-case input Lipschitz constant, and hence to large worst-case sensitivity. 
The mechanism is tied to the isotropizing effect of normalization layers. Batch normalization tends to push representations toward a more isotropic geometry~\citep{daneshmand2021batch}, but near rank-deficiency this same effect can amplify small perturbations into large output separations.

\paragraph{Smooth optimization.}
Dimension-free convergence guarantees have a long history in smooth optimization. For smooth nonconvex objectives, gradient descent can find an approximate first-order stationary point at a rate that does not depend on the ambient dimension~\citep{nesterov2013introductory,ghadimi2013stochastic}. A large body of work has tried to go beyond first-order stationarity and reach second-order stationary points, where the gradient is small and there are no directions of significant negative curvature~\citep{ge2015escaping,jin2017escape,jin2021nonconvex}. Under smoothness assumptions, such guarantees are possible in polynomial time. However, dimension-free second-order guarantees cannot hold in full generality. \citet{simchowitz2017gap} show that some dependence on the dimension is unavoidable for general smooth functions. Motivated by this limitation, \citet{daneshmand2018escaping} identify noise conditions under which stochastic gradient methods can still achieve dimension-free convergence to second-order stationary points, using structure that is natural in deep learning losses. These results rely on smoothness of the objective, an assumption that is often violated in modern neural networks because of ReLU activations, and max-pooling operations. For this reason, we focus here on nonsmooth optimization, where the objective need not be differentiable everywhere.

\paragraph{Nonsmooth optimization.}
The study of stationarity for nonsmooth Lipschitz functions goes back to the work of~\citet{goldstein1977optimization}, which introduced what are now known as Goldstein stationary points. Unlike in the smooth setting, exact first-order stationarity is not a finite-time target in general nonsmooth nonconvex optimization. This has led recent work to focus on relaxed notions such as $(\delta,\epsilon)$-Goldstein stationarity. In this direction, \citet{zhang2020complexity} gave one of the first non-asymptotic algorithms for finding such points, with complexity scaling polynomially in the Lipschitz constant $L$, the radius parameter $1/\delta$, and the accuracy $1/\epsilon$. More recently, \citet{davis2022gradient} developed a gradient-sampling approach under a standard oracle model, improving the dependence on $\epsilon$ and $L$ in certain regimes, at the cost of an additional dependence on the dimension. While these guarantees are polynomial for a fixed Lipschitz constant $L$, this distinction is important in deep learning, where $L$ is not an external constant. It is induced by the network architecture and can grow with depth. Our results show that, for deep networks with normalization layers, this growth can be exponential. Consequently, nonsmooth optimization guarantees that are polynomial in $L$ can become exponential in depth, even when they are formally dimension-free.

\section{Proofs and Details for \Cref{sec:exp_lower}} \label{appex:proof_lip_exp}

We begin with a proof sketch of \Cref{thm:main}, and then give the full proof of \Cref{thm:main,thm:param_lip} together with the required machinery and lemmas.

\subsection{Proof Sketch of \Cref{thm:main}} \label{sec:sketch}

We specialize to the square orthogonal regime~\eqref{eq:square_setting} and the recursion~\eqref{eq:network_recursion}.
Since $W_\ell\in\mathbb{O}(k)$, the linear maps are non-expansive, so any growth must come from the
normalization. Our argument exploits a tension between (i) the ability of $\Bn(\cdot)$ to
\emph{regularize} anisotropy when combined with orthogonal mixing, and (ii) the instability of this
regularization near rank-deficient inputs.

We follow \citet{joudaki2023impact} and track isotropy via the functional $I(\cdot)$ in
Definition~\ref{def:isotropy}. For normalized matrices $X=\Bn(\cdot)$ we have
$\tr(X^\top X)=k$ by~\eqref{eq:fro_norm_fixed}, hence $I(X)=\detm(X^\top X)^{1/k}\in(0,1]$.
A key input is the one-step lift inequality of \citet{meterez2024towards}:
Theorem~\ref{thm:onestep} shows that for $X=\Bn(\cdot)$ and Haar-uniform $W\in\mathbb{O}(k)$,
\[
\mathbb{E}\!\left[I(\Bn(XW))\,\middle|\,X\right]\ \ge\ \frac{1}{1-S(X)}\,I(X),
\]
where $S(\cdot)$ is the spectral deviation from Definition~\ref{def:spec_dev}.
Lemma~\ref{lem:uniformgap} provides a uniform gap $S(X)\ge s_{k,\eta}>0$ whenever $I(X)\le 1-\eta$.
Combining these yields a deterministic multiplicative lift: Corollary~\ref{cor:lift} guarantees that,
as long as $I(X)\le 1-\eta$, we can choose $W$ so that $I(\Bn(XW))\ge r_{k,\eta} I(X)$ with
$r_{k,\eta}=(1-s_{k,\eta})^{-1}>1$. Iterating this choice gives the greedy geometric increase
Corollary~\ref{cor:geom}, so after $m$ layers we can enforce $I(X_m)\ge 1-\eta$ provided the initial
isotropy is at least $r_{k,\eta}^{-m}(1-\eta)$.

We realize such an initialization while keeping it exponentially close to rank deficiency.
As in the appendix proof of Theorem~\ref{thm:main}, we set
$G_\varepsilon=(1-\varepsilon)\one\one^\top+\varepsilon\Id$ and choose $X_0$ with $X_0^\top X_0=G_\varepsilon$.
Then $\diag(G_\varepsilon)=\Id$, so $X_0=\Bn(X_0)$, and $I(X_0)=\detm(G_\varepsilon)^{1/k}\ge \varepsilon^{(k-1)/k}$.
Choosing
\[
\varepsilon \;=\; (1-\eta)^{\frac{k}{k-1}}\,r_{k,\eta}^{-\frac{mk}{k-1}}
\]
ensures $r_{k,\eta}^m I(X_0)\ge 1-\eta$, hence (by Corollary~\ref{cor:geom}) we obtain $I(X_m)\ge 1-\eta$ along the
resulting trajectory. Near-isotropy then forces uniform conditioning: Lemma~\ref{lem:lmin} yields
$\lambda_{\min}(X_m^\top X_m)\ge c_k I(X_m)^k \ge c_k(1-\eta)^k=:\delta_{k,\eta}$.

To turn this into a Lipschitz lower bound, we compare $X_0$ to a rank-$(k-1)$ neighbor.
Let $X_0=U\Sigma V^\top$ be an SVD with singular values $\sigma_1\ge\cdots\ge\sigma_k>0$ and define the matrix 
$Y_0:=U\diag(\sigma_1,\dots,\sigma_{k-1},0)V^\top$. Then $\rank(Y_0)<k$ and
$\|X_0-Y_0\|_F=\sigma_k=\sqrt{\varepsilon}$. Let $Y_t:=F_t(Y_0)$ under the same weights (one may, if desired,
interpret $\Bn$ via the standard $\varepsilon$-regularization from Remark~\ref{remark_esp_zero} to avoid degenerate columns; this does not
change the bound). Since each layer is an invertible right-multiplication followed by column-wise rescaling,
rank cannot increase, so $\rank(Y_t)<k$ for all $t$. Applying Lemma~\ref{lem:dist} at depth $m$ gives
$\|X_m-Y_m\|_F^2\ge \lambda_{\min}(X_m^\top X_m)\ge \delta_{k,\eta}$, and therefore
\[
\Lip_{\mathcal{X}}(F_m)\ \ge\ \frac{\|F_m(X_0)-F_m(Y_0)\|_F}{\|X_0-Y_0\|_F}
\ \ge\ \sqrt{\frac{\delta_{k,\eta}}{\varepsilon}}
\ =\ C_{k,\eta}\,r_{k,\eta}^{\frac{mk}{2(k-1)}}
\ =\ C_{k,\eta}e^{\gamma_{k,\eta}m},
\]
with $\gamma_{k,\eta}=\tfrac{k}{2(k-1)}\log r_{k,\eta}$ and $C_{k,\eta}$ as in Theorem~\ref{thm:main}.
Finally, the orthogonal equivariance $\Bn((QX)W)=Q\Bn(XW)$ for $Q\in\mathbb{O}(k)$ implies
$F_m(QX)=QF_m(X)$, so the same lower bound is witnessed by the uncountable family of pairs $(QX_0,QY_0)$,
as recorded at the end of the proof of Theorem~\ref{thm:main}.

\subsection{Isotropy and a One-step Lift Inequality } 

We use the notion of isotropy from \citet{joudaki2023impact}\footnote{In \citet{joudaki2023impact}, this quantity is referred to as \emph{isometry}; here, we use the term \emph{isotropy} for terminological clarity and to better reflect its role in our analysis of Lipschitz constants.}, together with the one-step lift developed in \citet{meterez2024towards}, to study the dynamics of linear neural networks with normalization across layers. Let $X\in\R^{k\times k}$ and assume $X^\top X\succ 0$.

\begin{definition}[Isotropy functional]\label{def:isotropy}
Define
\[
I(X)\coloneqq \frac{\detm(X^\top X)^{1/k}}{\tr(X^\top X)/k}\in(0,1].
\]
\end{definition}

The quantity $I(X)$ is scale-invariant and satisfies $I(X)=1$ if and only if
$X^\top X$ is a scalar multiple of the identity. In this case, we say that we have ``perfect isotropy''.
If $X=\Bn(\cdot)$ (so that $\tr(X^\top X)=k$), then $I(X)=\detm(X^\top X)^{1/k}$.

\begin{definition}[Spectral deviation]\label{def:spec_dev}
Suppose $X=\Bn(\cdot)$ and let $\lambda_1,\dots,\lambda_k$ be the eigenvalues of $X^\top X$
(so $\sum_{i=1}^k\lambda_i=\tr(X^\top X)=k$).
Define
\[
S(X)\coloneqq \frac{1}{2k^2(k+2)}\sum_{i=1}^k(\lambda_i-1)^2.
\]
\end{definition}

\begin{theorem}[One-step isotropy lift {\cite[Theorem~A.4]{meterez2024towards}}]\label{thm:onestep}
Let $X=\Bn(\cdot)$ and let $W$ be Haar-uniform on $\mathbb{O}(k)$.
Then
\[
\mathbb{E}_W\!\big[I(\Bn(XW))\,\big|\,X\big]\;\ge\;\frac{1}{1-S(X)}\,I(X).
\]
\end{theorem}

\subsection{A Uniform Multiplicative Lift Below an Isotropy Cap}

Fix $\eta\in(0,1)$ and define the compact set of eigenvalue vectors
\[
K_{\eta}\coloneqq \Bigl\{\lambda\in[0,\infty)^k:\ \sum_{i=1}^k\lambda_i=k,\ \Bigl(\prod_{i=1}^k\lambda_i\Bigr)^{1/k}\le 1-\eta\Bigr\}.
\]
Consider $S(\lambda)\coloneqq \frac{1}{2k^2(k+2)}\sum_{i=1}^k(\lambda_i-1)^2$ from Definition~\ref{def:spec_dev}.

We now prove a uniform spectral gap away from $I(X) = 1$.

\begin{lemma}\label{lem:uniformgap}
Let $\eta\in(0,1)$ and set $s_{k,\eta}\coloneqq \min_{\lambda\in K_\eta} S(\lambda)$.
Then $s_{k,\eta}>0$.
Moreover, if $X=\Bn(\cdot)$ and $I(X)\le 1-\eta$, then $S(X)\ge s_{k,\eta}$.
\end{lemma}

\begin{proof}
$K_\eta$ is nonempty, closed, and bounded, hence compact, and $S$ is continuous, so the minimum is
attained.
Also $S(\lambda)=0$ iff $\lambda_1=\cdots=\lambda_k=1$, but that vector is not in $K_\eta$ since its
geometric mean is $1>1-\eta$. Thus $s_{k,\eta}>0$.
The final statement follows because $S(X)=S(\lambda(X^\top X))$ and, for $X=\Bn(\cdot)$,
$I(X)=\bigl(\prod_{i=1}^k\lambda_i\bigr)^{1/k}$.
\end{proof}

In turn, we can convert this result into a deterministic multiplicative lift for the isotropy functional $I(X)$, as long as it is below $1 - \eta$.

\begin{corollary}\label{cor:lift}
Fix $\eta\in(0,1)$ and define
\[
r_{k,\eta}\coloneqq \frac{1}{1-s_{k,\eta}}>1.
\]
If $X=\Bn(\cdot)$ and $I(X)\le 1-\eta$, then there exists an orthogonal matrix $W\in\mathbb{O}(k)$ such that
\[
I(\Bn(XW))\ge r_{k,\eta}\,I(X).
\]
\end{corollary}

\begin{proof}
By Lemma~\ref{lem:uniformgap}, $S(X)\ge s_{k,\eta}$.
Theorem~\ref{thm:onestep} then gives
$\mathbb{E}_W[I(\Bn(XW))\mid X]\ge r_{k,\eta}I(X)$.
Since $I(\Bn(XW))\ge 0$, there must exist a realization of $W$ achieving at least the expectation.
\end{proof}

Consequently, we can show that the isotropy functional $I(X)$ increases geometrically until $X$ reaches near-isotropy. In the proof, we choose the weight matrices greedily according to \Cref{cor:lift}.

\begin{corollary}\label{cor:geom}
Fix $\eta\in(0,1)$ and let $r_{k,\eta}>1$ be as above.
Let $X_0=\Bn(\cdot)$ and choose $W_1,\dots,W_m\in\mathbb{O}(k)$ greedily as follows: for each $t\ge 1$,
if $I(X_{t-1})\le 1-\eta$ choose $W_t$ satisfying Corollary~\ref{cor:lift}, and set
$X_t\coloneqq \Bn(X_{t-1}W_t)$.
Then for every $t\ge 0$,
\[
I(X_t)\;\ge\; \min\bigl\{r_{k,\eta}^{\,t}\,I(X_0),\ 1-\eta\bigr\}.
\]
\end{corollary}

\begin{proof}
If $I(X_{t-1})\le 1-\eta$, then by construction $I(X_t)\ge r_{k,\eta}I(X_{t-1})$.
Once $I(X_{t-1})>1-\eta$ we do not require any further increase.
An induction yields the claimed bound.
\end{proof}

\subsection{Near-isotropy Forces a Floor on $\lambda_{\min}$}

A useful basic result is the inequality that connects the minimum eigenvalue in the spectrum of $X^\top X$ to the isotropy functional $I(X)$.

\begin{lemma}\label{lem:lmin}
Let $X=\Bn(\cdot)$ and let $\lambda_1\ge\cdots\ge \lambda_k$ be the eigenvalues of $X^\top X$
(so $\sum_{i=1}^k\lambda_i=k$ and $\prod_{i=1}^k\lambda_i=I(X)^k$).
Then
\[
\lambda_{\min}(X^\top X)=\lambda_k \;\ge\; c_k\, I(X)^k,
\qquad
c_k\coloneqq \Bigl(\frac{k-1}{k}\Bigr)^{k-1}.
\]
\end{lemma}

\begin{proof}
Set $\beta\coloneqq \lambda_k$ and note $\sum_{i=1}^{k-1}\lambda_i=k-\beta$.
By AM--GM, for fixed sum $k-\beta$ the product of the first $k-1$ eigenvalues is maximized when they
are equal, hence
\[
\prod_{i=1}^{k-1}\lambda_i \;\le\;\Bigl(\frac{k-\beta}{k-1}\Bigr)^{k-1}\;\le\;\Bigl(\frac{k}{k-1}\Bigr)^{k-1}.
\]
Therefore
\[
I(X)^k=\prod_{i=1}^k\lambda_i = \beta\prod_{i=1}^{k-1}\lambda_i
\;\le\; \beta \Bigl(\frac{k}{k-1}\Bigr)^{k-1},
\]
which rearranges to $\beta\ge ((k-1)/k)^{k-1} I(X)^k$.
\end{proof}

\subsection{Distance to a Rank-deficient Trajectory}

We next use a simple geometric fact that turns the rank mismatch between the two trajectories into a quantitative separation. In our construction, one trajectory becomes well-conditioned, while the other remains rank-deficient throughout the network. The following lemma shows that any rank-deficient matrix must stay at least a smallest-singular-value distance away from a well-conditioned full-rank matrix.

\begin{lemma}\label{lem:dist}
Let $X,Y\in\R^{k\times k}$ and suppose that $\operatorname{rank}(Y)<k$.
Then
\[
\|X-Y\|_F^2\;\ge\;\lambda_{\min}(X^\top X).
\]
\end{lemma}

\begin{proof}
Since $\operatorname{rank}(Y)<k$ there exists a unit vector $v\in\ker(Y)$.
Then
\[
\|X-Y\|_F^2=\tr\bigl((X-Y)^\top(X-Y)\bigr)
\ge v^\top (X-Y)^\top(X-Y) v
= \|(X-Y)v\|_2^2 = \|Xv\|_2^2.
\]
Finally, $\|Xv\|_2^2 = v^\top X^\top X v\ge \lambda_{\min}(X^\top X)$.
\end{proof}

\subsection{Exponential growth of $\Lip_{\mathcal{X}}(F_m)$ with depth}
\label{app:proof_main}

\begin{repeatthm}{thm:main}
    [Exponential lower bound on $\Lip_{\mathcal{X}}(F_m)$]
Fix $k\ge 2$ and constant $\eta\in(0,1)$. There exist constants
$C_{k,\eta}>0$ and $\gamma_{k,\eta}>0$ such that, for every depth
$m\in\mathbb{N}$, one can choose orthogonal weights
$W_1,\dots,W_m\in\mathbb{O}(k)$ with $\|W_\ell\|_2=1$ for all
$\ell\in[m]$, for which the depth-$m$ map $F_m$ defined
in~\eqref{eq:network_recursion}--\eqref{eq:square_setting} satisfies
\[
\Lip_{\mathcal{X}}(F_m)\;\ge\; C_{k,\eta}\,e^{\gamma_{k,\eta} m},
\]
where constants $C_{k,\eta}$ and $\gamma_{k,\eta}$ depend on $k,\eta$. Moreover, for this same fixed choice of weights, the lower bound is
witnessed by an uncountable family of input pairs.
\end{repeatthm}
\begin{proof}
Fix $m\in\mathbb{N}$ and let $r\coloneqq r_{k,\eta}>1$ from Corollary~\ref{cor:lift}.
Define
\[
\delta_{k,\eta}\coloneqq c_k(1-\eta)^k,
\qquad
\gamma_{k,\eta}\coloneqq \frac{k}{2(k-1)}\log r,
\qquad
C_{k,\eta}\coloneqq \sqrt{\delta_{k,\eta}}\,(1-\eta)^{-\frac{k}{2(k-1)}}.
\]

\paragraph{Step 1: choose an initialization $X_0$ with controlled anisotropy.}
Let
\[
G_\varepsilon \coloneqq (1-\varepsilon)\,\one\one^\top + \varepsilon\,\Id\in\R^{k\times k},
\qquad \varepsilon\in(0,1].
\]
Then $\diag(G_\varepsilon)=\Id$ and the eigenvalues of $G_\varepsilon$ are
$\lambda_1=k-(k-1)\varepsilon$ and $\lambda_2=\cdots=\lambda_k=\varepsilon$.
Choose any $X_0\in\R^{k\times k}$ such that $X_0^\top X_0=G_\varepsilon$
(e.g.\ take $X_0=L^\top$ where $G_\varepsilon=LL^\top$ is a Cholesky factorization).
Since $\diag(G_\varepsilon)=\Id$, every column of $X_0$ has norm $1$, hence
$X_0=\Bn(X_0)$.

Moreover,
\[
I(X_0)=\detm(G_\varepsilon)^{1/k}
=\bigl(\varepsilon^{k-1}(k-(k-1)\varepsilon)\bigr)^{1/k}
\;\ge\;\varepsilon^{\frac{k-1}{k}},
\]
because $k-(k-1)\varepsilon\ge 1$ for $\varepsilon\in(0,1]$.

\paragraph{Step 2: pick $\varepsilon$ so that $m$ greedy steps reach near-isotropy.}
Choose
\[
\varepsilon \coloneqq (1-\eta)^{\frac{k}{k-1}}\, r^{-\frac{mk}{k-1}}.
\]
Then $r^m I(X_0)\ge r^m\varepsilon^{(k-1)/k}=1-\eta$.

Now choose $W_1,\dots,W_m$ greedily as in Corollary~\ref{cor:geom} and set
$X_t\coloneqq \Bn(X_{t-1} W_t)$.
By Corollary~\ref{cor:geom}, $I(X_m)\ge 1-\eta$.

\paragraph{Step 3: eigenvalue floor after reaching near-isotropy.}
Lemma~\ref{lem:lmin} yields
\[
\lambda_{\min}(X_m^\top X_m)\;\ge\; c_k\,I(X_m)^k\;\ge\;c_k(1-\eta)^k=\delta_{k,\eta}.
\]

\paragraph{Step 4: a rank-deficient companion $Y_0$ and distance growth.}
Let $X_0=U\Sigma V^\top$ be an SVD with singular values $\sigma_1\ge\cdots\ge \sigma_k>0$.
Define the rank-$(k-1)$ matrix
\[
Y_0\coloneqq U\diag(\sigma_1,\dots,\sigma_{k-1},0)V^\top.
\]
Then $\operatorname{rank}(Y_0)<k$ and
\[
\|X_0-Y_0\|_F=\sigma_k=\sqrt{\lambda_{\min}(X_0^\top X_0)}=\sqrt{\varepsilon}.
\]

Let $Y_t\coloneqq F_t(Y_0)$.
Since each layer is multiplication by an invertible matrix ($W_t$) followed by multiplication by a
diagonal matrix (coming from $\Bn$), the rank cannot increase, hence $\operatorname{rank}(Y_t)<k$ for all $t$.

Applying Lemma~\ref{lem:dist} at depth $m$ gives
\[
\|X_m-Y_m\|_F^2\ge \lambda_{\min}(X_m^\top X_m)\ge \delta_{k,\eta}.
\]
Therefore
\[
\Lip_{\mathcal{X}}(F_m)\;\ge\;\frac{\|F_m(X_0)-F_m(Y_0)\|_F}{\|X_0-Y_0\|_F}
\;\ge\;\sqrt{\frac{\delta_{k,\eta}}{\varepsilon}}
=
C_{k,\eta}\,r^{\frac{mk}{2(k-1)}}
=
C_{k,\eta}\,e^{\gamma_{k,\eta}m}.
\]

\paragraph{Uncountably many witness pairs for the same weights.}
For any orthogonal $Q\in\mathbb{O}(k)$, we have for all $X$ and all orthogonal $W$,
\[
\Bn((QX)W)=Q\,\Bn(XW),
\]
because $((QX)W)^\top((QX)W)=W^\top X^\top Q^\top Q X W=W^\top X^\top X W$.
Hence $F_m(QX)=QF_m(X)$ by induction.
Consequently, the ratio
$\|F_m(QX_0)-F_m(QY_0)\|_F/\|QX_0-QY_0\|_F$ equals the ratio for $(X_0,Y_0)$ for every $Q\in\mathbb{O}(k)$.
Since $\mathbb{O}(k)$ is uncountable, this yields uncountably many witness pairs.
\end{proof}

\begin{repeatcorollary}{cor:parambudget}
Fix $\eta\in(0,1)$ and set $k=2$.
There exist constants $C_\eta>0$ and $\Gamma_\eta>0$ such that, for every integer $d\ge 4$, there is a depth-$m$ network with at most $d$ scalar weight parameters whose input--output map satisfies
\[
\Lip_{\mathcal{X}}(F_m)\;\ge\; C_\eta\,e^{\Gamma_\eta d}.
\]
\end{repeatcorollary}
\begin{proof}
When $k=2$, each layer contains $4$ scalar parameters if the weight matrix is counted as an unconstrained $2\times 2$ matrix. Thus, a depth-$m$ network has $4m$ scalar parameters. Given a budget $d\ge 4$, choose $m=\lfloor d/4\rfloor$, so that $4m\le d$. Applying \Cref{thm:main} with $k=2$ gives
\[
\Lip_{\mathcal{X}}(F_m)
\ge C_{2,\eta}e^{\gamma_{2,\eta}\lfloor d/4\rfloor}
\ge C_{2,\eta}e^{-\gamma_{2,\eta}} e^{(\gamma_{2,\eta}/4)d}.
\]
The claim follows by setting
\[
C_\eta \coloneqq C_{2,\eta}e^{-\gamma_{2,\eta}},
\qquad
\Gamma_\eta \coloneqq \gamma_{2,\eta}/4.
\]
\end{proof}

\subsection{Exponential growth of $\Lip_{\Theta}(\mathcal{F}_m)$ with depth} \label{app:exp_wrt_param}

\begin{repeatthm}{thm:param_lip}
    [Exponential lower bound on $\Lip_{\Theta}(\mathcal{F}_m)$]
Fix $k\ge 2$ and constant $\eta\in(0,1)$. There exist constants
$\widetilde C_{k,\eta}>0$ and $\widetilde\gamma_{k,\eta}>0$ depending on $k$ and $\eta$ such that, for every depth $m\ge 2$, there is a depth-$m$ network of the form~\eqref{eq:network_recursion}, with
$\|W_\ell\|_2\le 1$ for all $\ell\in[m]$, for which the fixed-input parameter-to-output map $
\mathcal{F}_m(\theta)\coloneqq F_m^\theta(X_{\mathrm{in}})
$
with $X_{\mathrm{in}}=\Id_k$
satisfies
\[
\Lip_{\Theta}(\mathcal{F}_m)
\;\ge\;
\widetilde C_{k,\eta}\,e^{\widetilde\gamma_{k,\eta}m}.
\]
\end{repeatthm}
\begin{proof}
Apply Theorem~\ref{thm:main} with depth $m-1$ to obtain orthogonal matrices
$\widetilde W_2,\dots,\widetilde W_m\in\mathbb{O}(k)$ and matrices $X_0,Y_0\in\mathbb{R}^{k\times k}$
with $\rank(Y_0)<k$ such that the corresponding depth-$(m-1)$ map
\[
G_{m-1}(Z)\coloneqq \Bn(\cdots \Bn(\Bn(Z\widetilde W_2)\widetilde W_3)\cdots \widetilde W_m)
\]
satisfies
\[
\|G_{m-1}(X_0)-G_{m-1}(Y_0)\|_F \ \ge\ \sqrt{\delta_{k,\eta}},
\qquad\text{and}\qquad
\|X_0-Y_0\|_F=\sqrt{\varepsilon},
\]
where $\varepsilon$ is exponentially small in $m$ as in the proof of Theorem~\ref{thm:main}.

Now consider a depth-$m$ network with input $X_{\mathrm{in}}=\Id_k$ and weights
\[
W_1 := \alpha X_0,\qquad W_1' := \alpha Y_0,\qquad W_\ell := \widetilde W_\ell\ \ (\ell=2,\dots,m),
\]
where $\alpha>0$ is chosen so that $\|W_1\|_2,\|W_1'\|_2\le 1$.
Since column-wise batch normalization is scale-invariant,
\[
\Bn(\alpha Z)=\alpha Z\,(D(\alpha Z))^{-1/2}=\alpha Z\,( \alpha^2 D(Z))^{-1/2}=\Bn(Z),
\]
we have $\Bn(X_{\mathrm{in}}W_1)=\Bn(X_0)$ and $\Bn(X_{\mathrm{in}}W_1')=\Bn(Y_0)$.
Moreover, $\rank(\Bn(Y_0))<k$.

Let $\theta=(W_1,\dots,W_m)$ and $\theta'=(W_1',W_2,\dots,W_m)$.
Then
\[
\mathcal{F}_m(\theta)=G_{m-1}(\Bn(X_0))
\qquad\text{and}\qquad
\mathcal{F}_m(\theta')=G_{m-1}(\Bn(Y_0)).
\]
The second trajectory remains rank-deficient at every depth, so Lemma~\ref{lem:dist} (applied at depth $m-1$)
and the conditioning lower bound from the proof of Theorem~\ref{thm:main} yield
\[
\|\mathcal{F}_m(\theta)-\mathcal{F}_m(\theta')\|_F \ \ge\ \sqrt{\delta_{k,\eta}}.
\]
On the other hand,
\[
\|\theta-\theta'\|_{2}=\|W_1-W_1'\|_F=\alpha\|X_0-Y_0\|_F=\alpha\sqrt{\varepsilon}.
\]
Therefore
\[
\Lip_{\Theta}(\mathcal{F}_m)\ \ge\
\frac{\|\mathcal{F}_m(\theta)-\mathcal{F}_m(\theta')\|_F}{\|\theta-\theta'\|_{2}}
\ \ge\ \frac{\sqrt{\delta_{k,\eta}}}{\alpha\sqrt{\varepsilon}}
\ =\ \widetilde C_{k,\eta}e^{\widetilde\gamma_{k,\eta} m},
\]
since $\varepsilon$ is exponentially small in $m$ (same as in Theorem~\ref{thm:main}).
\end{proof}

\begin{repeatcorollary}{cor:objective_param_lip}
Let $\mathcal{F}_m$ be the parameter-to-output map constructed in \Cref{thm:param_lip}. There exists a realizable target parameter $\theta^\star\in\Theta$ such that the objective $f$ defined in~\eqref{eq:def_objective} satisfies
\[
\Lip(f)
\;\ge\;
\widetilde C_\eta e^{\widetilde \Gamma_\eta d},
\]
up to the constants in \Cref{cor:parambudget_weights}.
\end{repeatcorollary}

\begin{proof}
In the construction of \Cref{thm:param_lip}, let $\theta^\star$ be one of the two parameter choices witnessing the lower bound on $\Lip_{\Theta}(\mathcal{F}_m)$, and let $\theta$ be the other one. Since the target is realizable, $f(\theta^\star)=0$. Therefore,
\[
    \Lip(f) \ge \frac{| f(\theta) - f(\theta^\star) |}{\| \theta - \theta^\star \|_2}
    = \frac{| f(\theta) |}{\| \theta - \theta^\star \|_2}
    = \frac{\|\mathcal{F}_m(\theta)-\mathcal{F}_m(\theta^\star)\|_F}{\| \theta - \theta^\star \|_2}.
\]
The right-hand side is exactly the same separation ratio used in the proof of \Cref{thm:param_lip}. Applying \Cref{cor:parambudget_weights} gives the claimed exponential lower bound.
\end{proof}

\subsection{Extension to Normalization with Mean Subtraction}
\label{app:centering}

The results of \Cref{sec:main_lip_input,sec:main_lip_param} extend to
batch normalization with mean subtraction. Under our notation in \Cref{sec:exp_lower}, the centered normalization map is
defined, for $X\in\R^{n\times k}$, by
\[
\Bn_{\mathrm{ctr}}(X)
\coloneqq
\Bn\!\left(
\left(\Id_n-\frac{1}{n}\one\one^\top\right)X
\right),
\]
whenever every centered column is nonzero. This corresponds to the
standard batch-variance normalization with a fixed affine scale of
$1/\sqrt n$. 

We now try to embed our hard instances into the centered subspace. Suppose $n\ge k+1$, and choose $U\in\R^{n\times k}$ satisfying
\[
U^\top U=\Id_k,
\qquad
\one^\top U=0.
\]
Such a matrix exists since the orthogonal complement of $\one$ in
$\R^n$ has dimension $n-1\ge k$. Left multiplication by $U$ is an
isometry and maps every matrix into the subspace of column-centered
matrices. In particular, for every $Z\in\R^{k\times k}$,
\[
(UZ)^\top(UZ)=Z^\top Z,
\]
so the embedding preserves rank, Gram eigenvalues, column norms,
and Frobenius distances.

Moreover, since $\one^\top U=0$, mean subtraction leaves $UZW$
unchanged. Consequently, for every $Z$ and $W$ for which normalization
is defined,
\[
\Bn_{\mathrm{ctr}}(UZW)
=
UZW D(ZW)^{-1/2}
=
U\Bn(ZW).
\numberthis{eq:centered_intertwining}
\]
Thus, centered normalization reproduces the original normalization
dynamics exactly on the embedded subspace.

Let $\overline F_m$ denote the depth-$m$ network obtained by replacing
each normalization layer of $F_m$ with $\Bn_{\mathrm{ctr}}$, while
keeping the same weight matrices. Applying
\eqref{eq:centered_intertwining} inductively yields
\[
\overline F_m(UZ)=UF_m(Z).
\]
In particular, applying this identity to the witness pair
$(X_0,Y_0)$ from \Cref{thm:main} gives
\[
\frac{
\|\overline F_m(UX_0)-\overline F_m(UY_0)\|_F
}{
\|UX_0-UY_0\|_F
}
=
\frac{
\|F_m(X_0)-F_m(Y_0)\|_F
}{
\|X_0-Y_0\|_F
}.
\]
Hence the exponential input Lipschitz lower bound carries over
with exactly the same initial perturbation size and final output
separation. The Gram-eigenvalue bounds underlying the construction
are also preserved by the embedding.

The parameter Lipschitz lower bound follows from the same identity.
Indeed, replacing the fixed input $\Id_k$ in \Cref{thm:param_lip}
with $U$, and using the same two weight tuples, preserves both
their parameter distance and the corresponding output separation.
Since the weight matrices remain $k\times k$, the parameter count
is unchanged.

Finally, the robustness construction of \Cref{thm:adv_radius}
extends directly to the centered network. Using the readout
$UA_m$ and the same offset $b_m$, we have $\|UA_m\|_F=1$ and
\[
\langle UA_m,\overline F_m(UZ)\rangle_F+b_m
=
\langle A_m,F_m(Z)\rangle_F+b_m.
\]
Thus, the embedded clean and adversarial inputs achieve exactly
the same classification margins and perturbation distance as in
\Cref{thm:adv_radius}. In particular, the exponential input and
parameter sensitivity bounds, as well as the exponentially small
adversarial radius, hold for mean-subtracted batch normalization.

\subsection{Extension to ReLU Activations}
\label{app:relu}

Our construction in \Cref{app:centering} also extends to networks with centered
batch normalization followed by ReLU activations. We use the
standard affine shift of batch normalization, fixed to $2$ in
every channel, and define
\[
\widetilde f_W(H)
\coloneqq
\operatorname{ReLU}\!\left(
\Bn_{\mathrm{ctr}}(HW)+2\one\one^\top
\right),
\]
where $\operatorname{ReLU}$ acts entrywise and
$\one\one^\top\in\R^{n\times k}$ denotes the all-ones matrix.
Due to normalization layers, each column of
$\Bn_{\mathrm{ctr}}(HW)$ has unit Euclidean norm, so every entry
lies in $[-1,1]$. Consequently, the affine shift makes every
entry strictly positive, and ReLU acts as the identity.

The key observation is that mean subtraction eliminates any
matrix with identical rows, even after multiplication by a
weight matrix. Combining this property with
\eqref{eq:centered_intertwining}, we obtain
\[
\begin{aligned}
\widetilde f_W(UZ)
&=U f_W(Z)+2\one\one^\top,\\
\widetilde f_W(UZ+2\one\one^\top)
&=U f_W(Z)+2\one\one^\top.
\end{aligned}
\]
Thus, the affine shift introduced by one layer is removed by
the centering operation in the next layer, while the centered
component follows exactly the dynamics of the original network.

Writing
$\widetilde F_m=\widetilde f_{W_m}\circ\cdots\circ
\widetilde f_{W_1}$, induction yields
\[
\widetilde F_m(UZ)
=
UF_m(Z)+2\one\one^\top.
\]
In particular, for the witness pair $(X_0,Y_0)$,
\[
\widetilde F_m(UX_0)-\widetilde F_m(UY_0)
=
U\bigl(F_m(X_0)-F_m(Y_0)\bigr).
\]
Since $U^\top U=\Id_k$, the input and output distances are
preserved exactly. The exponential input Lipschitz lower bound
therefore holds with the same witness ratio. Likewise, using
the fixed input $U$ and the weight tuples from
\Cref{thm:param_lip} preserves their parameter distance and
output separation, yielding the same parameter Lipschitz
lower bound. All affine shifts are fixed, so the number of
free weight parameters is unchanged.

Finally, the robustness construction transfers under the
same readout $UA_m$ and offset $b_m$. Indeed, since
$\one^\top U=0$,
\[
\begin{aligned}
\langle UA_m,\widetilde F_m(UZ)\rangle_F+b_m
&=
\langle UA_m,UF_m(Z)+2\one\one^\top\rangle_F+b_m\\
&=
\langle A_m,F_m(Z)\rangle_F+b_m.
\end{aligned}
\]
Hence the clean and adversarial inputs keep the same score
margins and perturbation distance as in
\Cref{thm:adv_radius}.

We conclude that the exponential input and parameter
Lipschitz lower bounds, together with the exponentially
small adversarial radius, are also attained by networks with
centered batch normalization and ReLU activations.

\subsection{Embedding the Construction at Larger Widths}
\label{app:width_embedding}

We show that the exponential lower bounds can be achieved with
constants independent of width by embedding the width-two construction
into networks of arbitrary width $k\ge 2$.

Let $X_0,Y_0$ and $W_1,\ldots,W_m$ be the witnesses from
\Cref{thm:main} for $k=2$. For any $k\ge 2$, consider the
block-diagonal matrices
\[
\diag(X_0,\Id_{k-2}),\qquad
\diag(Y_0,\Id_{k-2}),\qquad
\diag(W_\ell,\Id_{k-2}),\quad \ell\in[m],
\]
where the identity blocks are omitted when $k=2$.
Since normalization acts columnwise and the additional columns
have unit norm, the two trajectories at depth $t$ are precisely
$\diag(X_t,\Id_{k-2})$ and $\diag(Y_t,\Id_{k-2})$.
Their input distance and final output separation are therefore
identical to those of the width-two construction. Consequently,
the input Lipschitz lower bound
\[
\Lip_X(F_m)\ge C_{2,\eta}e^{\gamma_{2,\eta}m}
\]
holds for every $k\ge 2$, with constants independent of width.

Similarly, applying the same embedding to the parameter witnesses
in \Cref{thm:param_lip}, with fixed input $\Id_k$, preserves their
parameter distance, output separation, and unit operator norms.
Thus, the parameter Lipschitz lower bound also holds uniformly
over width, with the constants of the width-two construction.

These bounds have width-independent exponential rates in depth.
When all entries of the weight matrices are counted, however,
the parameter dimension is $d=mk^2$, and the input lower bound
becomes
\[
\Lip_X(F_m)\ge
C_{2,\eta}\exp\!\left(\frac{\gamma_{2,\eta}}{k^2}d\right).
\]

\section{Proofs and Details for \Cref{sec:imp_opt}} \label{appex:imp_opt}

\subsection{Background on Goldstein Stationarity and First-order Oracle} \label{app:goldstein}

\paragraph{Notation and Background. } Let $f:\R^d\to\R$ be (locally) $L$-Lipschitz with respect to $\|\cdot\|_2$:
\[
|f(x)-f(y)|\le L\|x-y\|_2,\qquad \forall x,y\in\R^d.
\]
Let $\partial f(x)$ denote the \emph{Clarke subdifferential} at $x$, which coincides
with $\{\nabla f(x)\}$ when $f$ is differentiable at $x$ and agrees with the usual convex subdifferential
when $f$ is convex. The canonical solution concept studied in nonsmooth nonconvex optimization is that of a \emph{Clarke stationary point}~\citep{clarke1975generalized}. A point $x \in \mathbb{R}^d$ is \emph{$\varepsilon$-Clarke stationary} if $\dist(0,\partial f(x)) \le \varepsilon$, where $\dist(0,Q) \defeq \min_{g\in \partial f(x)}\|g\|_2$. 

\citet{zhang2020complexity} showed that for general nonconvex Lipschitz functions, Clarke stationarity cannot be guaranteed in a finite number of oracle queries in the worst case, even under very strong first-order oracle access. This result shows that the consideration of more relaxed solution concepts, in favor of tractability, is inevitable.

\paragraph{Goldstein Stationarity. } A classical and now-standard relaxation is due to \citet{goldstein1977optimization}. Fix $\delta>0$.
The key idea is to \emph{convexify the subdifferential over a neighborhood}:
\[
\partial_\delta f(x)
\;\coloneqq\;
\conv\Bigl(\,\bigcup_{y:\ \|y-x\|_2<\delta}\partial f(y)\Bigr).
\]
We say that $x$ is \emph{$(\delta,\varepsilon)$-Goldstein stationary} if
\[
\dist\bigl(0,\partial_\delta f(x)\bigr)\ \le\ \varepsilon.
\]
This definition recovers the smooth optimality condition when $f$ is differentiable and $\delta$ is sufficiently small. The recent breakthrough of \citet{zhang2020complexity} highlighted the tractability of this solution concept in nonsmooth settings, and subsequent works have further developed and studied it extensively~\citep{davis2022gradient,lin2022gradient,jordan2022complexity,jordan2023deterministic,kong2023cost,cutkosky2023optimal}.

\paragraph{Oracle model.}
We consider the standard first-order oracle model used in the nonsmooth optimization literature. At each query point $x_t$, the oracle reveals the function value $f(x_t)$ together with subgradient information at $x_t$ (e.g., a Clarke subgradient $g_t \in \partial f(x_t)$ or even the full set $\partial f(x_t)$). The lower bounds stated in the paper holds even when the oracle returns the entire set $\partial f(x_t)$.

\subsection{Proofs and Details for \Cref{sec:lower_bound_goldstein}}
\label{app:L_lower_bound}

\begin{theorem}[{\citep[Theorem~3]{kornowski2022complexity}}]
\label{thm:KS-eps-lb}
For any $\varepsilon<1$ and any $\delta\le \frac{1}{12\varepsilon}$, the following holds. 
For every randomized first-order algorithm $\mathcal{A}$, there exists a convex $1$-Lipschitz function 
$f:\mathbb{R}^d\to\mathbb{R}$, with $d=\widetilde{\mathcal{O}}(1/\varepsilon^6)$, such that 
$f(x_1)-\inf_x f(x)\le 1$, but
\[
\Pr_{\mathcal{A}}\Bigl[
\exists\, t\le T \text{ such that } x_t
\text{ is }(\delta,\varepsilon)\text{-Goldstein stationary}
\Bigr]
< \frac{1}{3},
\]
unless $T=\Omega(1/\varepsilon^2)$.
\end{theorem}

We now immediately show that no randomized algorithm can avoid a polynomial dependence on $L$.
\begin{repeatcorollary}{cor:polyL-lb}[Formal]
Fix $L>0$ and $\varepsilon\in(0,L)$, and suppose that
$\delta\le \frac{L}{12\varepsilon}$. Then, for any randomized first-order
algorithm $\mathcal{A}$, there exists a convex $L$-Lipschitz function
$F:\mathbb{R}^d\to\mathbb{R}$, with
$d=\widetilde{\mathcal{O}}((L/\varepsilon)^6)$, such that
$F(x_1)-\inf_x F(x)\le L$, and
\[
\Pr_{\mathcal{A}}\Bigl[
\exists\, t\le T \text{ such that } x_t
\text{ is }(\delta,\varepsilon)\text{-Goldstein stationary}
\Bigr]
< \frac{2}{3}
\]
unless
\[
T=\Omega\!\left(\frac{L^2}{\varepsilon^2}\right).
\]
In particular, in the worst case, any randomized first-order method needs
$\Omega(L^2/\varepsilon^2)$ oracle calls to find a
$(\delta,\varepsilon)$-Goldstein stationary point with probability at least
$2/3$.
\end{repeatcorollary}
\begin{proof}
Set $\varepsilon' \coloneqq \varepsilon/L$. Since $\varepsilon\in(0,L)$, we have
$\varepsilon'\in(0,1)$. Applying \Cref{thm:KS-eps-lb} with accuracy
$\varepsilon'$ gives a convex $1$-Lipschitz function
$f:\mathbb{R}^d\to\mathbb{R}$, with
$d=\widetilde{\mathcal{O}}(1/\varepsilon'^6)
=\widetilde{\mathcal{O}}((L/\varepsilon)^6)$, such that
$f(x_1)-\inf_x f(x)\le 1$, and no randomized first-order algorithm can find a
$(\delta,\varepsilon')$-Goldstein stationary point with constant probability
unless
\[
T=\Omega(1/\varepsilon'^2).
\]
The condition on $\delta$ becomes
\[
\delta \le \frac{1}{12\varepsilon'}
= \frac{L}{12\varepsilon}.
\]

Now define $F\coloneqq Lf$. Then $F$ is convex and $L$-Lipschitz, and
\[
F(x_1)-\inf_x F(x)
=
L\bigl(f(x_1)-\inf_x f(x)\bigr)
\le L.
\]
Moreover, by positive homogeneity of the Clarke and Goldstein subdifferentials,
\[
\partial_\delta F(x)=L\,\partial_\delta f(x),
\]
and therefore
\[
\operatorname{dist}(0,\partial_\delta F(x))
=
L\,\operatorname{dist}(0,\partial_\delta f(x)).
\]
Thus, $x$ is $(\delta,\varepsilon)$-Goldstein stationary for $F$ if and only if
it is $(\delta,\varepsilon/L)$-Goldstein stationary for $f$, i.e.,
$(\delta,\varepsilon')$-Goldstein stationary for $f$.

Consequently, any randomized first-order algorithm that finds a
$(\delta,\varepsilon)$-Goldstein stationary point of $F$ with probability at
least $2/3$ would also find a $(\delta,\varepsilon')$-Goldstein stationary point
of $f$ with the same probability. By \Cref{thm:KS-eps-lb}, this requires
\[
T=\Omega(1/\varepsilon'^2)
=
\Omega\!\left(\frac{L^2}{\varepsilon^2}\right).
\]
This proves the claim.
\end{proof}

\begin{remark}
The regime in \Cref{cor:polyL-lb} is compatible with our exploding-Lipschitz setting. After rescaling the lower bound from \citep{kornowski2022complexity}, the hard instance has dimension $\widetilde{\mathcal{O}}((L/\varepsilon)^6)$ and requires $\delta \le L/(12\varepsilon)$. In our deep normalized networks, $L$ can be $\Omega(\exp(d))$, so these conditions are mild for fixed or polynomially scaled accuracy parameters. Thus the lower-bound regime overlaps with the regime arising in our construction, and substituting $L=\Omega(\exp(d))$ into $\Omega(L^2/\varepsilon^2)$ yields exponential dependence on the network dimension.
\end{remark}

\paragraph{Why exploding $L$ undermines ``dimension-free'' rates.}
The discussion in \Cref{sec:exp_lower} shows that the Lipschitz constant cannot be treated as a harmless constant in deep normalized networks. Existing guarantees for finding $(\delta,\varepsilon)$-Goldstein stationary points scale polynomially in the objective Lipschitz constant $L$, and \Cref{cor:polyL-lb} shows that some polynomial dependence on $L$ is unavoidable in the worst case. But in our setting, $L$ is induced by the network itself. By \Cref{thm:param_lip}, it can grow exponentially with depth, and hence with the parameter dimension. Therefore, a guarantee of the form $\poly(L,1/\delta,1/\varepsilon)$ may be formally independent of dimension, but it can still translate into exponential dependence on the dimension for deep normalized networks.

\begin{algorithm}[!t]
\small
\caption{Probabilistic Greedy Goldstein Descent (PGGD)}
\label{alg:pgd}
\begin{algorithmic}[1]
\Require $x_0\in\mathbb{R}^d$, $\delta>0$, $\tau>0$, $\rho\in(0,1)$, $\gamma\in(0,1)$, $f_{\mathrm{lb}}\le \inf f$.

\State $\Delta \gets f(x_0)-f_{\mathrm{lb}}$
\State $T_{\max}\gets \left\lceil \Delta/(\tau\delta)\right\rceil$
\State $m \gets \left\lceil \frac{1}{\rho}\log\!\frac{T_{\max}+1}{\gamma}\right\rceil$
\State $x\gets x_0$

\For{$t=0$ \textbf{to} $T_{\max}$}
    \State Sample i.i.d.\ $v_1,\dots,v_m\sim \mathrm{Unif}(\mathbb{S}^{d-1})$
    \State Query $f(x-\delta v_i)$ for all $i$ and let $i^\star\in\arg\min_i f(x-\delta v_i)$

    \If{$f(x-\delta v_{i^\star}) \le f(x)-\tau\delta$}
        \State $x \gets x-\delta v_{i^\star}$
    \Else
        \State \Return $x$
    \EndIf
\EndFor

\State \Return $x$
\end{algorithmic}
\end{algorithm}

\subsection{Proofs and Details for \Cref{sec:quest}}
\label{app:pgs}

For fixed $x$ and $\delta>0$, recall
\[
m_\delta(x;v)\coloneqq
\min_{g\in\partial_\delta f(x)}\langle g,v\rangle,
\qquad v\in\mathbb{S}^{d-1}.
\]
The following elementary identity is the dual form of Goldstein stationarity.

\begin{lemma}\label{lem:minimax_phi}
For any $\varepsilon\ge 0$,
\[
\dist(0,\partial_\delta f(x))\le \varepsilon
\qquad\Longleftrightarrow\qquad
\max_{v\in\mathbb{S}^{d-1}} m_\delta(x;v)\le \varepsilon .
\]
Moreover, if $\dist(0,\partial_\delta f(x))>0$, then
\[
\dist(0,\partial_\delta f(x))
=
\max_{v\in\mathbb{S}^{d-1}} m_\delta(x;v).
\]
\end{lemma}

\begin{proof}
Let $Q=\partial_\delta f(x)$ and let $g^\star$ be the projection of $0$ onto $Q$.
For any unit vector $v$,
\[
m_\delta(x;v)\le \langle g^\star,v\rangle\le \|g^\star\|_2.
\]
Thus $\max_v m_\delta(x;v)\le \dist(0,Q)$. If $g^\star\neq 0$, take
$v^\star=g^\star/\|g^\star\|_2$. The projection optimality condition gives
$\langle g-g^\star,g^\star\rangle\ge 0$ for all $g\in Q$, hence
\[
\langle g,v^\star\rangle\ge \|g^\star\|_2
\qquad\text{for all }g\in Q.
\]
Therefore $m_\delta(x;v^\star)=\|g^\star\|_2=\dist(0,Q)$. If $g^\star=0$, then
$\dist(0,Q)=0$, and the stated equivalence for $\varepsilon\ge 0$ follows immediately.
\end{proof}

\begin{lemma}
\label{lem:margin_implies_decrease}
Let $f:\R^d\to\R$ be Lipschitz. Fix $x\in\R^d$ and $\delta>0$.
There exists a full-measure set $\mathcal{V}_x\subset\mathbb{S}^{d-1}$ such that, for every
$v\in\mathcal{V}_x$ and every $\tau\in\R$,
\[
m_\delta(x;v)\ge \tau
\qquad\Longrightarrow\qquad
f(x-\delta v)\le f(x)-\tau\delta .
\]
\end{lemma}

\begin{proof}
Fix $x$. By \citet[Lemma~2.3]{davis2022gradient}, there is a set
$\mathcal{V}_x\subset\mathbb{S}^{d-1}$ of full surface measure such that, for every
$v\in\mathcal{V}_x$, the function $t\mapsto f(x-tv)$ is differentiable for a.e.\
$t\in[0,\delta]$ and satisfies
\[
f(x)-f(x-\delta v)
=
\int_0^\delta \langle \nabla f(x-tv),v\rangle\,dt .
\]
Now suppose $m_\delta(x;v)\ge \tau$. For a.e.\ $t\in[0,\delta]$, whenever
$f$ is differentiable at $x-tv$, we have
\[
\nabla f(x-tv)\in \partial f(x-tv)\subseteq \partial_\delta f(x),
\]
since $x-tv\in B_\delta(x)$ for all $t<\delta$ and the endpoint is irrelevant for the integral.
Therefore,
\[
\langle \nabla f(x-tv),v\rangle \ge \tau
\qquad\text{for a.e. }t\in[0,\delta].
\]
Integrating gives
\[
f(x)-f(x-\delta v)
\ge
\int_0^\delta \tau\,dt
=
\tau\delta,
\]
which is equivalent to the desired inequality.
\end{proof}

\begin{repeatthm}{thm:pgd}
\Cref{alg:pgd} finds a $(\delta,\tau,\rho)$-PGS point with probability at least $1-\gamma$ using $
\mathcal{O}\!\left(
\frac{\Delta}{\tau\delta\rho}
\log\frac{\Delta}{\gamma\tau\delta}
\right) $
queries. In particular, for $(\tau,\rho)=(\varepsilon,\varepsilon)$, this becomes $
\mathcal{O}\!\left(
\frac{\Delta}{\delta\varepsilon^2}
\log\frac{\Delta}{\gamma\delta\varepsilon}
\right)$, with no polynomial dependence on $L$.
\end{repeatthm}
\begin{proof}
Each accepted step decreases $f$ by at least $\tau\delta$. Since
$f(x_0)-f_{\rm lb}=\Delta$, the algorithm can accept at most
$T_{\max}\le \Delta/(\tau\delta)$ steps, and hence performs at most
$T_{\max}+1$ stopping checks.

Call a point $x$ bad if
\[
\Pr_{V\sim\Unif(\mathbb{S}^{d-1})}
\big[f(x-\delta V)\le f(x)-\tau\delta\big]>\rho.
\]
At any bad point, the probability that all $m$ sampled directions fail to give sufficient decrease is at most
\[
(1-\rho)^m\le e^{-\rho m}\le \frac{\gamma}{T_{\max}+1}.
\]
A union bound over the stopping checks shows that, with probability at least
$1-\gamma$, the algorithm never stops at a bad point. Thus the returned point
$x_{\mathrm{out}}$ satisfies
\[
{\Pr}_{V}
\big[f(x_{\mathrm{out}}-\delta V)\le f(x_{\mathrm{out}})-\tau\delta\big]\le \rho.
\]

By Lemma~\ref{lem:margin_implies_decrease}, for almost every $V$,
\[
m_\delta(x_{\mathrm{out}};V)\ge \tau
\quad\Longrightarrow\quad
f(x_{\mathrm{out}}-\delta V)\le f(x_{\mathrm{out}})-\tau\delta.
\]
Therefore,
\[
{\Pr}_V \big[m_\delta(x_{\mathrm{out}};V)\ge \tau\big]\le \rho,
\]
so $x_{\mathrm{out}}$ is $(\delta,\tau,\rho)$-PGS. The query bound follows from
at most $T_{\max}+1$ iterations and $m+1$ function evaluations per iteration.
\end{proof}

\section{Proofs and Details for \Cref{sec:imp_gene_robust}}
\label{app:adv_robustness}

This appendix makes the two informal robustness statements from \Cref{sec:imp_gene_robust} precise. We reuse the rank-separated witness pair constructed in the proof of \Cref{thm:main}.

\begin{repeatthm}{thm:adv_radius}
Fix $k\ge 2$ and $\eta\in(0,1)$. There exist constants $a_{k,\eta}>0$, $\gamma_{k,\eta}>0$, and $\mu_{k,\eta}>0$ such that, for every depth $m\in\mathbb{N}$, one can choose orthogonal weights $W_1,\ldots,W_m\in\mathbb{O}(k)$ and inputs $X_{\mathrm{clean}},X_{\mathrm{adv}}\in\R^{k\times k}$ with the following properties:
\[
\rank(X_{\mathrm{clean}})=k-1,
\qquad
\rank(X_{\mathrm{adv}})=k,
\qquad
\|X_{\mathrm{adv}}-X_{\mathrm{clean}}\|_F\le a_{k,\eta}e^{-\gamma_{k,\eta}m}.
\]
Moreover, there is a linear readout $A_m\in\R^{k\times k}$ and an offset $b_m\in\R$ with $\|A_m\|_F=1$ such that the scalar score
\[
s_m(Z)\coloneqq \langle A_m,F_m(Z)\rangle_F+b_m
\]
satisfies
\[
s_m(X_{\mathrm{clean}})\ge \mu_{k,\eta},
\qquad
s_m(X_{\mathrm{adv}})\le -\mu_{k,\eta}.
\]
Consequently, for the binary classifier $Z\mapsto \operatorname{sign}(s_m(Z))$ with clean label $+1$ at $X_{\mathrm{clean}}$,
\[
\inf\bigl\{\|E\|_F:\; s_m(X_{\mathrm{clean}}+E)\le 0\bigr\}
\le a_{k,\eta}e^{-\gamma_{k,\eta}m}.
\]
\end{repeatthm}

\begin{proof}
Use the construction in the proof of \Cref{thm:main}. With $r=r_{k,\eta}>1$ from \Cref{cor:lift}, that proof constructs inputs $Y_0$ and $X_0$ such that
\[
\rank(Y_0)=k-1,
\qquad
\rank(X_0)=k,
\qquad
\|X_0-Y_0\|_F\le 2(1-\eta)^{\frac{k}{2(k-1)}}r^{-\frac{mk}{2(k-1)}}.
\]
It also constructs orthogonal weights for which, writing $Y_m=F_m(Y_0)$ and $X_m=F_m(X_0)$,
\[
\|X_m-Y_m\|_F^2\ge \delta_{k,\eta},
\qquad
\delta_{k,\eta}:=c_k(1-\eta)^k,
\]
where $c_k=((k-1)/k)^{k-1}$ is the constant from \Cref{lem:lmin}.
Set
\[
X_{\mathrm{clean}}:=Y_0,
\qquad
X_{\mathrm{adv}}:=X_0,
\qquad
\gamma_{k,\eta}:=\frac{k}{2(k-1)}\log r,
\qquad
 a_{k,\eta}:=2(1-\eta)^{\frac{k}{2(k-1)}}.
\]
Then $\|X_{\mathrm{adv}}-X_{\mathrm{clean}}\|_F\le a_{k,\eta}e^{-\gamma_{k,\eta}m}$.

It remains to choose the readout. Let
\[
\Delta_m:=\|Y_m-X_m\|_F\ge \sqrt{\delta_{k,\eta}},
\qquad
A_m:=\frac{Y_m-X_m}{\Delta_m},
\qquad
b_m:=-\frac12\langle A_m,Y_m+X_m\rangle_F.
\]
Then $\|A_m\|_F=1$, and direct calculation gives
\[
s_m(X_{\mathrm{clean}})=\langle A_m,Y_m\rangle_F+b_m=\frac{\Delta_m}{2},
\qquad
s_m(X_{\mathrm{adv}})=\langle A_m,X_m\rangle_F+b_m=-\frac{\Delta_m}{2}.
\]
Thus both inequalities hold with
\[
\mu_{k,\eta}:=\frac12\sqrt{\delta_{k,\eta}}.
\]
Finally, the perturbation $E=X_{\mathrm{adv}}-X_{\mathrm{clean}}$ is feasible for the displayed infimum because $s_m(X_{\mathrm{clean}}+E)=s_m(X_{\mathrm{adv}})\le0$. This proves the adversarial-radius bound.
\end{proof}

\begin{repeatprop}{prop:rank_creating}
In the construction of \Cref{thm:adv_radius}, the perturbation $X_{\mathrm{adv}}-X_{\mathrm{clean}}$ is rank-creating:
\[
\rank(X_{\mathrm{clean}})=k-1,
\qquad
\rank(X_{\mathrm{adv}})=k.
\]
Furthermore, if $Z\in\R^{k\times k}$ is any rank-deficient input for which the forward pass is well-defined, then $F_m(Z)$ is rank-deficient and
\[
\|F_m(X_{\mathrm{adv}})-F_m(Z)\|_F
\ge \sqrt{\delta_{k,\eta}},
\qquad
\delta_{k,\eta}=c_k(1-\eta)^k.
\]
Thus the full-rank adversarial branch is separated by a constant from the entire rank-deficient output branch.
\end{repeatprop}

\begin{proof}
The rank identities for $X_{\mathrm{clean}}$ and $X_{\mathrm{adv}}$ were proved in \Cref{thm:adv_radius}. Now let $Z$ be rank-deficient and suppose the forward pass is well-defined. Each layer maps
\[
Z_{t-1}\longmapsto \Bn(Z_{t-1}W_t)=Z_{t-1}W_tD(Z_{t-1}W_t)^{-1/2}.
\]
Here $W_t$ is invertible and, because the normalization is well-defined, $D(Z_{t-1}W_t)^{-1/2}$ is an invertible diagonal matrix. Hence the rank cannot increase through a layer, so $F_m(Z)$ is rank-deficient.

On the full-rank branch of the construction, the proof of \Cref{thm:main} gives
\[
\lambda_{\min}(F_m(X_{\mathrm{adv}})^\top F_m(X_{\mathrm{adv}}))\ge \delta_{k,\eta}.
\]
Applying \Cref{lem:dist} with $X=F_m(X_{\mathrm{adv}})$ and $Y=F_m(Z)$ yields
\[
\|F_m(X_{\mathrm{adv}})-F_m(Z)\|_F^2
\ge \lambda_{\min}(F_m(X_{\mathrm{adv}})^\top F_m(X_{\mathrm{adv}}))
\ge \delta_{k,\eta},
\]
which proves the claim.
\end{proof}

\section{Experiments}
\label{app:exp}

\paragraph{Low-rank structure in image datasets.}

\begin{figure*}[!t]
    \centering

    \begin{subfigure}{0.95\textwidth}
        \centering
        \includegraphics[width=\linewidth]{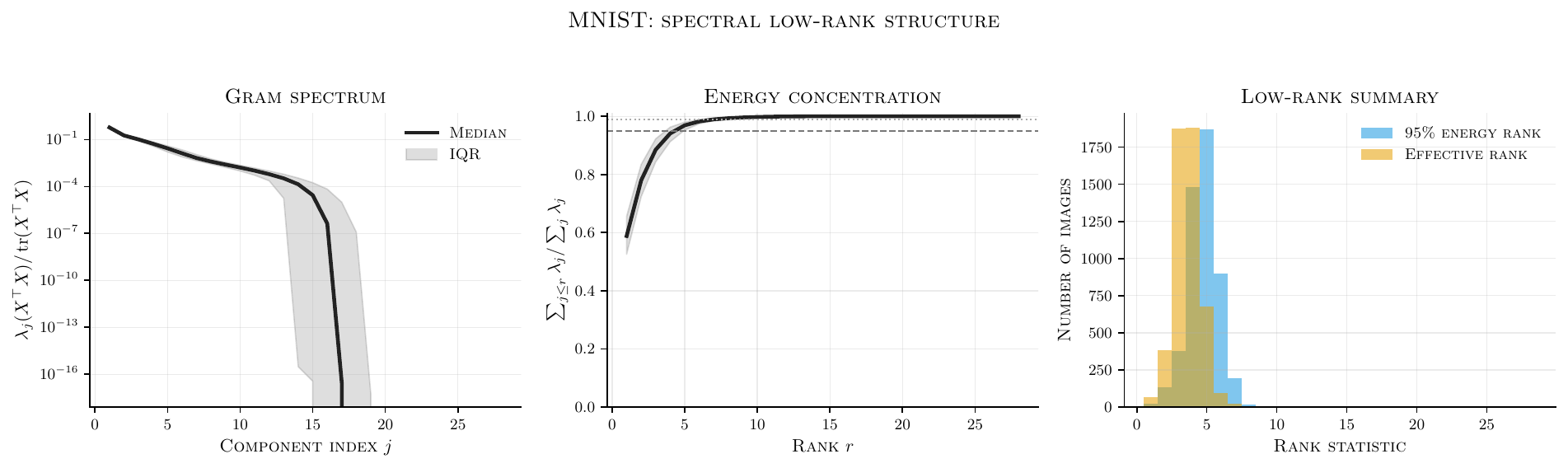}
        \label{fig:mnist_spectral_rank}
    \end{subfigure}

    \vspace{0.4cm}

    \begin{subfigure}{0.95\textwidth}
        \centering
        \includegraphics[width=\linewidth]{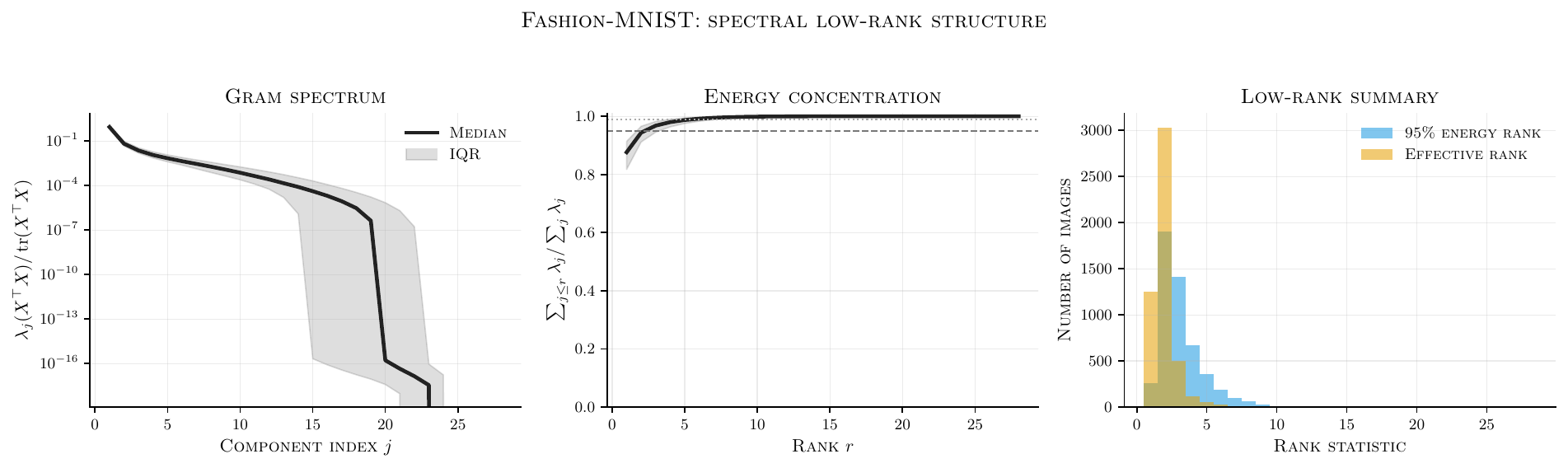}
        \label{fig:fashionmnist_spectral_rank}
    \end{subfigure}

    \caption{
    Spectral evidence for approximate low-rank structure in MNIST and Fashion-MNIST. 
    For each dataset, the left panel shows the median normalized Gram spectrum $\lambda_j(X^\top X)/\tr(X^\top X)$ with interquartile range, the middle panel shows cumulative spectral energy with the $90\%$ and $95\%$ thresholds, and the right panel reports the distributions of effective rank and the rank needed to capture $95\%$ of the energy. 
    In both datasets, most of the spectral energy is concentrated in a few components, indicating that typical images lie close to a low-rank set.
}
\label{fig:mnist_fashion_spectral_structure}
\end{figure*}

To support the rank-separation mechanism studied in the main text, we examine the spectral structure of standard image datasets. As shown in \Cref{fig:mnist_fashion_spectral_structure,fig:cifar10_spectral_summary}, images from MNIST, Fashion-MNIST, and CIFAR-10 exhibit strong approximate low-rank structure. Their spectra decay rapidly, and most of the spectral energy is concentrated in a small number of singular components. Thus, while the images are not necessarily exactly rank-deficient, they typically lie close to a low-rank set. This observation is consistent with classical low-rank models of image data~\citep{savas2007handwritten,zhang2013learning,gu2014weighted,xu2015patch,liu2016content} and motivates our focus on perturbations that move an input away from its low-rank structure.

\paragraph{Perturbation model for MNIST and Fashion-MNIST.}

In the MNIST and Fashion-MNIST experiments, we view each image as a matrix $X \in [0,1]^{28 \times 28}$ and compare several perturbation families at matched pre-clipping energy. Given a perturbation $\Delta$, the perturbed image is
\[
    \widetilde X = \Pi_{[0,1]}\!\left(X+\Delta\right),
\]
where $\Pi_{[0,1]}$ denotes entrywise clipping to the valid pixel range. The noise level is controlled by a variance parameter $\sigma^2$, and before clipping every sampled perturbation is rescaled so that
\[
    \|\Delta\|_F = \sigma \sqrt{28 \cdot 28}.
\]
Thus all noise families have the same root-mean-square pixel energy before clipping, making the curves directly comparable across perturbation types. We consider four perturbation structures. IID pixel noise samples a full-dimensional Gaussian direction and is full rank with probability one, so it directly tests rank-creating perturbations. Singular-value (or same-subspace) noise perturbs only the singular values of the image and therefore stays within the row and column subspaces already present in $X$. Column-offset and row-offset noise broadcast one Gaussian vector across rows or columns, respectively, producing structured rank-one perturbations. Together, these perturbation families allow us to separate the effect of perturbation magnitude from the effect of perturbation geometry.

\begin{figure*}[h]
    \centering

    \begin{subfigure}{0.95\textwidth}
        \centering
        \includegraphics[width=\linewidth]{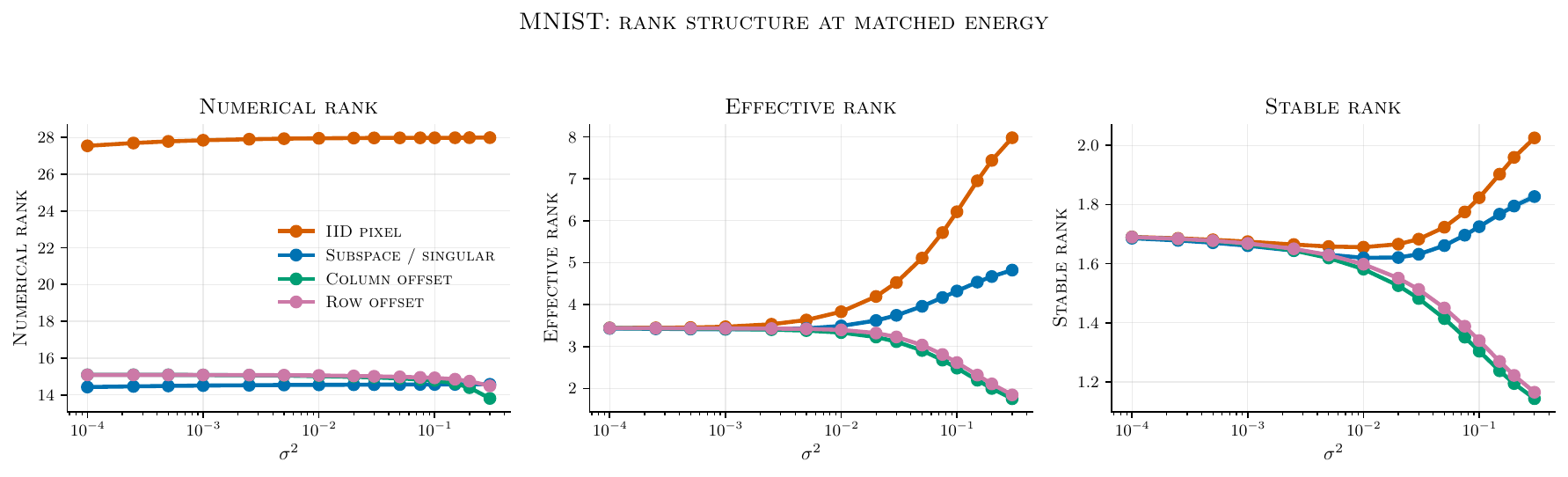}
        \label{fig:mnist_rank_inflation}
    \end{subfigure}

    \vspace{0.35cm}

    \begin{subfigure}{0.95\textwidth}
        \centering
        \includegraphics[width=\linewidth]{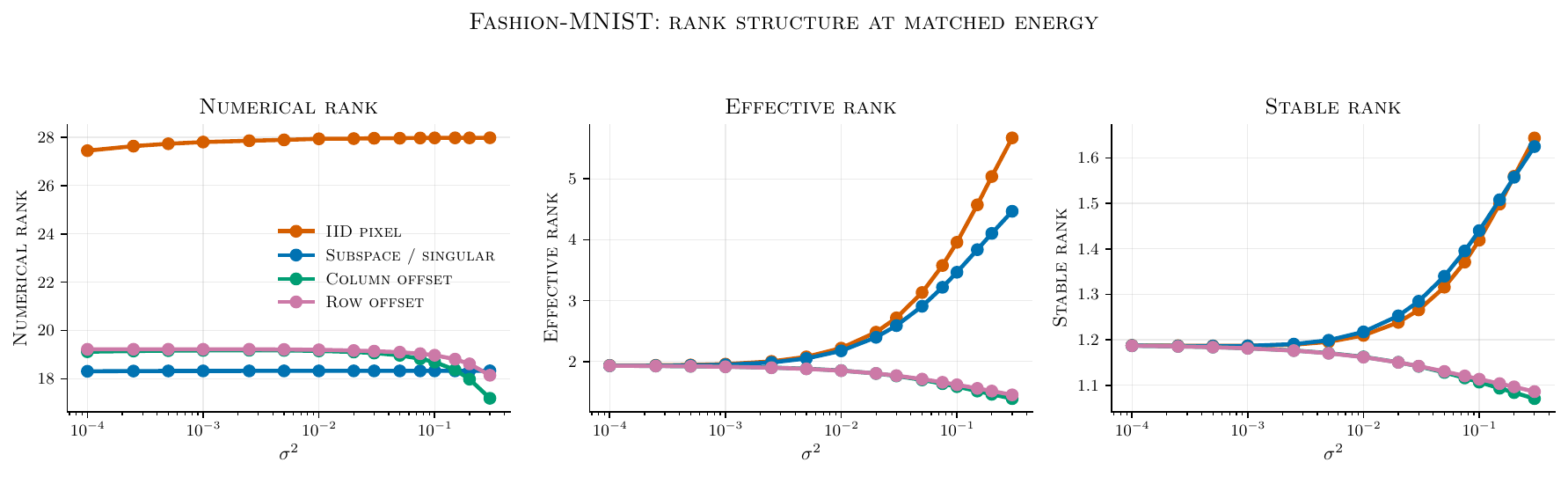}
        \label{fig:fashion_rank_inflation}
    \end{subfigure}

    \caption{
    Rank structure under matched-energy perturbations on MNIST and Fashion-MNIST. 
    For each dataset, we report numerical rank, effective rank, and stable rank as the perturbation variance $\sigma^2$ increases. 
    IID pixel noise strongly inflates the intrinsic rank measures, while same-subspace, column-wise, and row-wise perturbations produce much smaller changes despite having the same pre-clipping energy.
    }

    \label{fig:mnist_fashion_rank_inflation}
\end{figure*}

\paragraph{Rank effects of matched-energy perturbations.}

\begin{figure}[t]
    \centering
    \includegraphics[width=\textwidth]{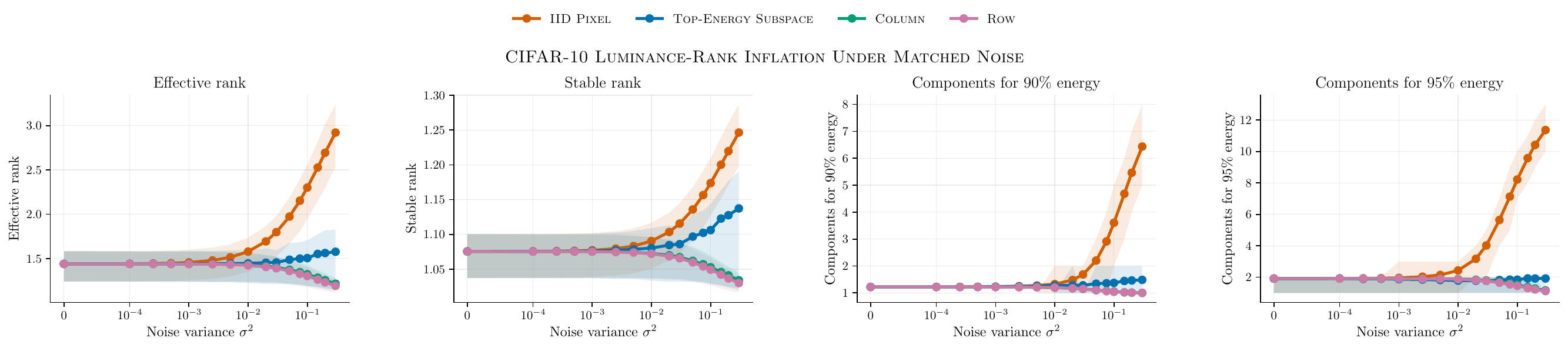}
    \caption{
    Rank inflation under matched-energy perturbations on CIFAR-10 luminance images. 
    The plots show effective rank, stable rank, and the number of components required to capture $90\%$ and $95\%$ of the spectral energy. 
    IID pixel noise substantially increases all rank-based measures as $\sigma^2$ grows, whereas same-subspace and structured row or column perturbations have a much milder effect.
    }
    \label{fig:cifar10_rank_inflation}
\end{figure}

We next examine how different perturbation geometries affect the intrinsic rank structure of the images when their pre-clipping energy is matched. As shown in \Cref{fig:mnist_fashion_rank_inflation,fig:cifar10_rank_inflation}, i.i.d. pixel noise consistently increases effective rank, stable rank, and the number of components needed to capture most of the spectral energy. In contrast, same-subspace perturbations have a much weaker effect, while row- and column-structured perturbations often preserve or even reduce these rank-based quantities. These observations support the mechanism in the main text that perturbations that create new spectral directions are more disruptive to the low-rank structure of images than perturbations that remain aligned with it.

\paragraph{Robustness under matched-energy perturbations.}

\begin{figure*}[t]
    \centering

    \begin{subfigure}{0.95\textwidth}
        \centering
        \includegraphics[width=\linewidth]{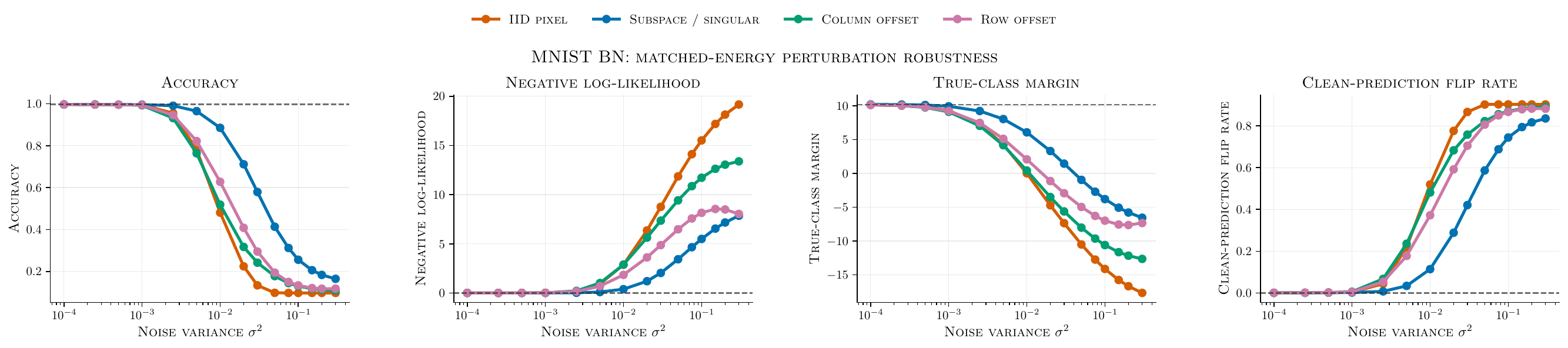}
        \label{fig:mnist_bn_noise}
    \end{subfigure}

    \vspace{0.6cm}

    \begin{subfigure}{0.95\textwidth}
        \centering
        \includegraphics[width=\linewidth]{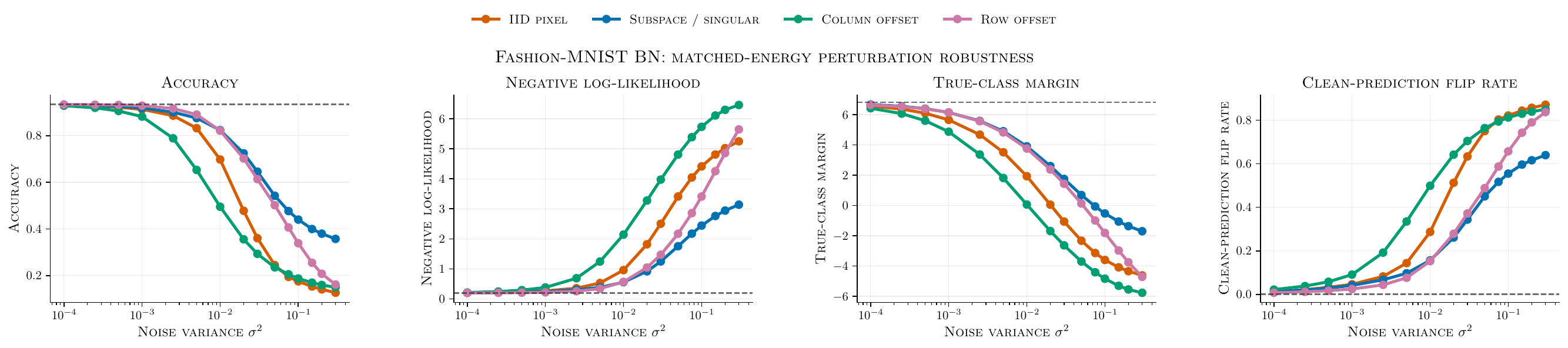}
        \label{fig:fashionmnist_bn_noise}
    \end{subfigure}

    \vspace{0.6cm}

    \begin{subfigure}{0.95\textwidth}
        \centering
        \includegraphics[width=\linewidth]{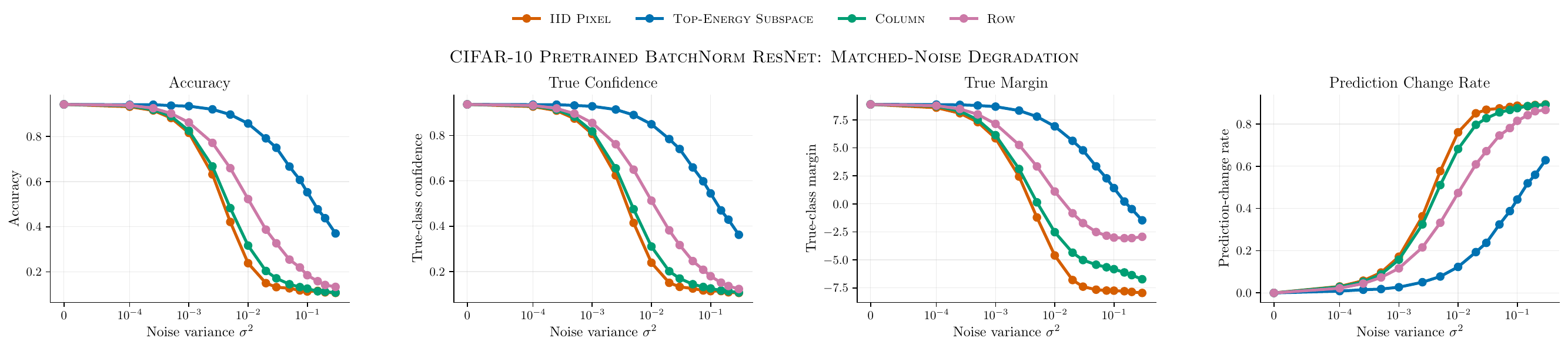}
        \label{fig:cifar10_bn_noise}
    \end{subfigure}

    \caption{
Matched-energy perturbation robustness across MNIST, Fashion-MNIST, and CIFAR-10 BatchNorm models. 
Each row shows performance as the noise variance $\sigma^2$ increases under i.i.d. pixel noise, subspace-preserving perturbations, column offsets, and row offsets. 
Subspace-preserving perturbations typically degrade accuracy, confidence, and margin more slowly and induce fewer prediction changes, while perturbations that move away from the image subspace cause earlier instability.
    }
\label{fig:matched_noise_degradation_all}
\end{figure*}

We next evaluate whether the geometric differences between perturbation families translate into different prediction behavior. \Cref{fig:matched_noise_degradation_all} shows that, even when perturbations are matched in pre-clipping energy, their effect on BatchNorm networks can differ substantially. Across MNIST, Fashion-MNIST, and CIFAR-10, subspace-preserving perturbations tend to preserve accuracy, confidence, and margin for larger noise levels, and they produce lower prediction-flip rates. In contrast, i.i.d. pixel noise and structured row or column perturbations often cause earlier confidence collapse, margin degradation, and prediction instability. This is consistent with the rank-separation mechanism from the main text: perturbations that move images away from their dominant low-rank structure are amplified more strongly than perturbations that remain aligned with that structure.

\begin{figure}[!h]
    \centering

        \begin{subfigure}{1\textwidth}
        \centering
        \includegraphics[width=\linewidth]{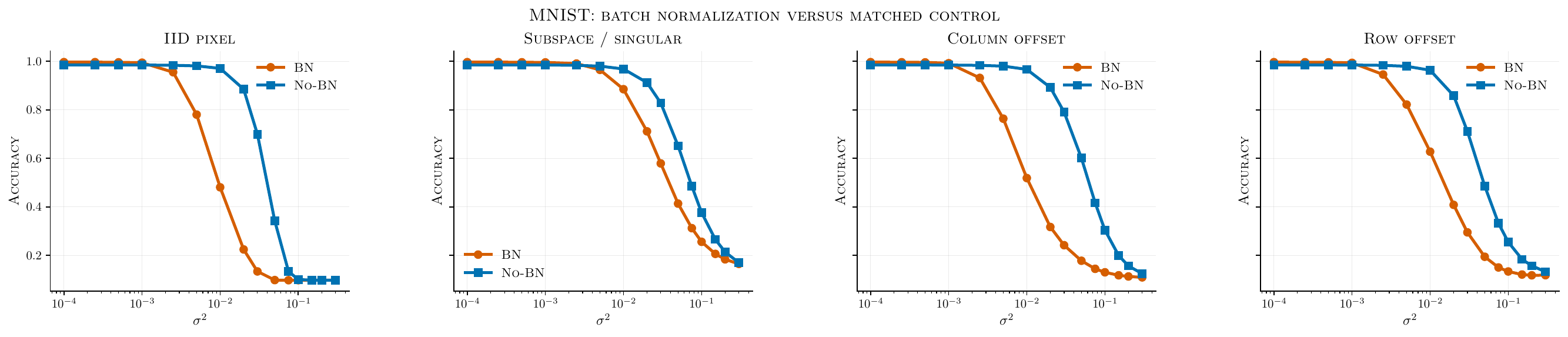}
        \label{fig:mnist_bn}
    \end{subfigure}

    \begin{subfigure}{1\textwidth}
        \centering
        \includegraphics[width=\linewidth]{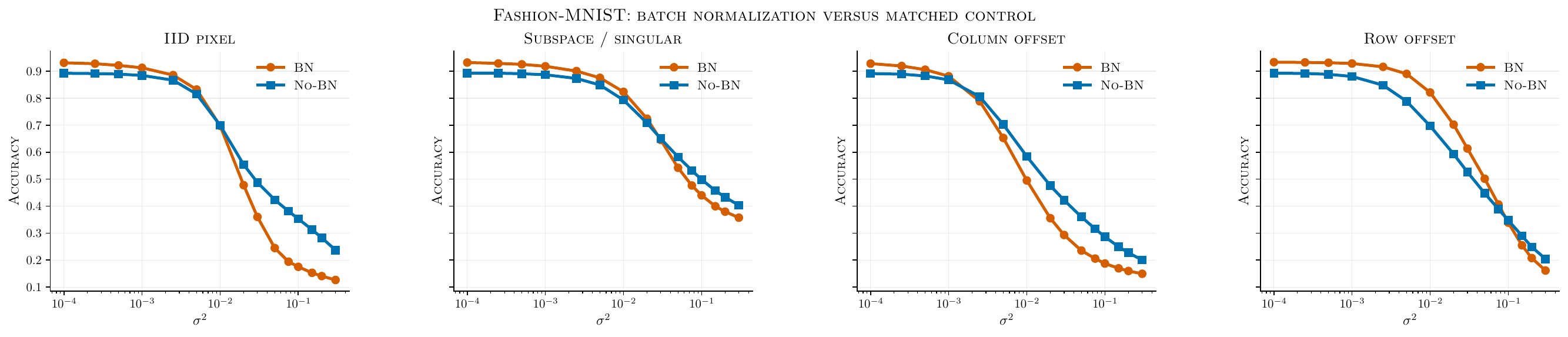}
        \label{fig:fashionmnist_bn}
    \end{subfigure}

    \vspace{0.5cm}

    \caption{
    Batch normalization versus matched NoBN controls under structured input perturbations. 
    Top row shows MNIST and bottom row shows Fashion-MNIST. 
    For each dataset, we compare accuracy as the perturbation variance $\sigma^2$ increases under i.i.d. pixel noise, same-subspace singular perturbations, column offsets, and row offsets. 
    BatchNorm models often remain accurate at low noise levels but exhibit sharper degradation at larger perturbation strengths, while matched NoBN controls typically degrade more gradually.
    }
    \label{fig:bn_vs_nobn_combined}
\end{figure}

\paragraph{Batch normalization versus matched controls.}

We next compare models with batch normalization to matched architectures without batch normalization under the same perturbation families. As shown in \Cref{fig:bn_vs_nobn_combined}, BatchNorm models often maintain high accuracy at very small noise levels but can undergo a much sharper accuracy drop once the perturbation variance crosses a moderate threshold. This effect is especially clear on MNIST under i.i.d. pixel noise, column offsets, and row offsets. In contrast, the matched NoBN models tend to degrade more gradually. Same-subspace perturbations remain comparatively less damaging, which is consistent with the rank-separation mechanism discussed in the main text.

\end{document}